  \documentclass[final,1p,times,twocolumn]{elsarticle}
\usepackage{amssymb}
\usepackage{amsmath}
\usepackage{amsthm}

\usepackage{enumerate} 
\usepackage{booktabs}
\usepackage{multirow} 
\usepackage{changes}
\usepackage{hyperref} 
\usepackage{algorithmic} 
\usepackage{algorithm} 

\newtheorem{definition}{Definition}
\newtheorem{lemma}{Lemma} 
\newtheorem{theorem}{Theorem}  
   
\newtheorem{corollary}{Corollary}

\journal{}
\begin{document}

\begin{frontmatter}

%% Title, authors and addresses

%% use the tnoteref command within \title for footnotes;
%% use the tnotetext command for theassociated footnote;
%% use the fnref command within \author or \affiliation for footnotes;
%% use the fntext command for theassociated footnote;
%% use the corref command within \author for corresponding author footnotes;
%% use the cortext command for theassociated footnote;
%% use the ead command for the email address,
%% and the form \ead[url] for the home page:
%% \title{Title\tnoteref{label1}}
%% \tnotetext[label1]{}
%% \author{Name\corref{cor1}\fnref{label2}}
%% \ead{email address}
%% \ead[url]{home page}
%% \fntext[label2]{}
%% \cortext[cor1]{}
%% \affiliation{organization={},
%%             addressline={},
%%             city={},
%%             postcode={},
%%             state={},
%%             country={}}
%% \fntext[label3]{}

\title{Fast Estimations of Lower Bounds on Hitting Time  of Elitist Evolutionary Algorithms}

\author[NTU]{Jun He\corref{cor1}}
\ead{jun.he@ntu.ac.uk}
\cortext[cor1]{Corresponding author}
\affiliation[NTU]{organization={Department of Computer Science, Nottingham Trent University},
             %city={Nottingham},
             postcode={NG11 8NS},
            country={U.K.}}
            
%% use optional labels to link authors explicitly to addresses:
%% \author[label1,label2]{}
%% \affiliation[label1]{organization={},
%%             addressline={},
%%             city={},
%%             postcode={},
%%             state={},
%%             country={}}
%%
%% \affiliation[label2]{organization={},
%%             addressline={},
%%             city={},
%%             postcode={},
%%             state={},
%%             country={}}

\author[SUST]{Siang Yew Chong} %% Author name
\ead{chongsy@sustech.edu.cn}
\affiliation[SUST]{organization={Department of Computer Science and Engineering, Southern University of Science and Technology },%Department and Organization
           % addressline={}, 
             city={Shenzhen },
            postcode={518055}, 
            %state={},
            country={China}}
%% Author affiliation
\author[LN]{Xin Yao} %% Author name
\ead{xinyao@ln.edu.hk}
  \affiliation[LN]{organization={Department of Computing and Decision Sciences, Lingnan University},
             %addressline={Tuen Mun},
             city={Hong Kong}
             %postcode={},
             %state={},
             %country={}
             } 

%% Abstract
\begin{abstract} 
The fitness level method is a widely used tool for estimating the hitting time of elitist evolutionary algorithms. This paper shows that when a fitness function does not exhibit a level‑based structure, the lower bounds obtained by partitioning the entire search space become inherently loose. To address this limitation, we introduce the subset fitness level method, which focuses on a selected subset of non‑optimal solutions, partitions this subset into fitness levels, and derives lower bounds on hitting time through hitting probability. We further develop explicit formulas, based on the notions of paths and segments, that enable fast computation of these bounds. The proposed method yields tight lower bounds for fitness functions lacking level‑based structure. 
\end{abstract}

%%Graphical abstract
%\begin{graphicalabstract}
%\includegraphics{grabs}
%\end{graphicalabstract}

%%Research highlights
\begin{highlights}
\item  A subset fitness level method is proposed to estimate lower bounds on the hitting time of elitist evolutionary algorithms.
\item This method can provide tight lower bounds for fitness functions that do not conform to  a level‑based structure. 
\end{highlights}

%% Keywords

\end{frontmatter}

%% main text

\section{Introduction}
\label{secIntroduction} Hitting time is the minimal number of generations that an evolutionary algorithm (EA) takes to find an optimal solution \citep{he2001drift,he2003towards}. The mean hitting time is an important indicator for evaluating the performance of EAs. An easy-to-use tool for estimating the hitting time of elitist EAs is the fitness level method~\citep{wegener2003methods,sudholt2012new,doerr2024lower}. 
Its working principle is to divide the entire search space into ranks $(S_0, \ldots, S_K)$ according to fitness values from high to low, where the highest rank $S_0$ represents the optimal solution set. Each rank is called a fitness level. Then, the hitting time is estimated using transition probabilities between fitness levels. Due to its ease of use, the fitness level method has been widely adopted for analyzing the time complexity of elitist EAs~\citep{aboutaib2022runtime,doerr2022towards,malalanirainy2022runtime,muhlenthaler2022exact,oliveto2022tight,friedrich2023analysis,hevia2023theoretical,lehre2023runtime,rajabi2023stagnation,shahrzad2023accelerating,zheng2023theoretical,cerf2024population,dang2024level,doerr2024runtime,jin2024set,krejca2024flexible,lengler2025runtime}.  The fitness level method is suitable for use to estimate an upper bound on the hitting time, but needs to be improved to estimate a lower bound \citep{doerr2024lower}. 

Recently, the fitness level method is formulated  as a special type of drift analysis with fitness level partitioning \citep{he2025drift,he2025estimation}. According to this theory~\cite{he2025estimation}, the lower bound on the mean hitting time from any   $X_k \in S_k$ to the optimal solution set $S_0$ (where $k=1, \ldots, K$) can be calculated from the transition probability and hitting probability: 
\begin{equation} 
\label{equ:Lower-bound}  \frac{1}{ \max_{X_k\in S_k}
p(X_k, S_0 \cup \cdots \cup S_{k-1})}+\sum^{k-1}_{\ell=1}\frac{ \min_{X_k \in S_k}
h(X_k, S_\ell)}{\max_{X_\ell \in S_\ell}
p(X_\ell, S_0 \cup \cdots \cup S_{\ell-1})}, 
\end{equation} 
where $p(X_\ell, S_0 \cup \cdots \cup S_{\ell-1})$ is the transition probability from $X_\ell \in S_\ell$ to the higher levels $S_0 \cup \cdots \cup S_{\ell-1}$ and  $h(X_k, S_\ell)$ is the hitting probability from $X_k$ to $S_\ell$. Similarly, the upper bound  on the mean hitting time from any   $X_k \in S_k$ to  $S_0$ is expressed as below \cite{he2025estimation}: 
\begin{equation} \label{equ:Upper-bound}  \frac{1}{ \min_{X_k\in S_k}
p(X_k, S_0 \cup \cdots \cup S_{k-1})}+\sum^{k-1}_{\ell=1}\frac{ \max_{X_k \in S_k}
h(X_k, S_\ell)}{\min_{X_\ell \in S_\ell}
p(X_\ell, S_0 \cup \cdots \cup S_{\ell-1})}, 
\end{equation} 

The calculation of the transition probability is straightforward. Therefore,  the problem of estimating the hitting time is transformed into estimating the hitting probability. According to the formulas \eqref{equ:Lower-bound} and \eqref{equ:Upper-bound}, given a lower bound $c_{k,\ell}$ for the hitting probability $\min_{X_k \in S_k}
h(X_k, S_\ell)$, a lower bound for the hitting time can be derived; given an upper bound $c_{k,\ell}$ for the hitting probability $\max_{X_k \in S_k}
h(X_k, S_\ell)$, an upper bound for the hitting time can be derived. These bounds $c_{k,\ell}$ are referred to as linear coefficients in \cite{he2025drift,he2025estimation} to distinguish them from the bounds of the hitting time. 
Early research~\cite{wegener2003methods} utilizes a trivial upper bound coefficient of $c_{k,\ell}=1$, which effectively produces a tight upper bound on the hitting time. Conversely, while setting the lower bound coefficient to $c_{k,\ell}=0$ is straightforward, this approach fails to yield a tight lower bound. A general method called drift analysis of hitting probability \cite{he2025estimation} is proposed to estimate the upper and lower bounds of the hitting probability. Furthermore, explicit expressions are constructed for calculating the hitting probability~\cite{he2025drift,he2025estimation}.

The mean hitting time to reach the optimal solution set is equal to the expected number of visits to all non-optimal solutions. A lower bound on the hitting time can be obtained by estimating the number of visits to some non-optimal solutions, but obtaining an upper bound requires considering the number of visits to all non-optimal solutions. Intuitively, obtaining a lower bound on the hitting time is faster than obtaining the upper bound. This paper proposes the subset fitness level method to derive lower bounds on the hitting time of elitist EAs. 

The subset method involves partitioning a subset of non-optimal solutions into fitness levels and calculating lower bound coefficients based on this refined level partition. New explicit formulas are proposed to estimate the hitting probability based on the concepts of paths and segments, and then lower bounds on the mean hitting time are quickly derived. This paper demonstrates two advantages of the subset method. First, it can quickly estimate the lower bound of the hitting time. Second, this method of partitioning a subset of non-optimal solutions into fitness levels yields a tighter lower bound than the traditional method that partitions the entire search space into fitness levels.

This paper is structured as follows.   Section~\ref{sec:Review} reviews the related work on lower bounds obtained from the fitness level method. Section~\ref{sec:Limitation} explains the limitation of the fitness level methods for lower bounds. Section~\ref{sec:Method} introduces the new subset fitness level method.  Section~\ref{sec:Applications} applies the new method to six knapsack instances. Section~\ref{sec:Conclusion} concludes the paper.

%%%%%%%%%%%%%%%%%%%%%%%%%%%%%%%%%%%%%%%%%%%%%%%%%%%%%%%%%%%%%%%%%
\section{Existing Lower Bounds Based on Fitness Levels}
\label{sec:Review} 
This section reviews existing lower bounds on the hitting time that are derived from the fitness level method. Before beginning the review, an overview of the relevant background concepts will be presented.

\subsection{Elitist EAs and Fitness Level Partitioning}
Consider an EA to maximize a fitness function $f({x})$, which is defined on a finite set. An EA is an iterative algorithm inspired by Darwinian evolution. It generates a sequence of solutions $(X ^{[t]})_{t \ge 0}$ where $X^{[t]}$ represents the solution(s) at the $t$th generation. The sequence  $(X ^{[t]})_{t \ge 0}$ can be modeled by a homogeneous Markov chain \citep{he2003towards}, where $X$ denotes a state of the Markov chain.  
The transition probability from $X$ to $Y$ is denoted by $p(X, Y)$.
Let $S$ denote the set of all states and $S_{\mathrm {opt}}$ denote the set of optimal states. For a population-based EA, $X$ denotes a population of solutions and its fitness is $f(X_{\mathrm{opt}}) =\max\{f(X), X \in S\}$.

The fitness level method is only applicable to elitist EAs, which satisfy \( f(X^{[t+1]}) \ge f(X^{[t]}) \) for any \( t \ge 0 \). The fitness level method is based on fitness level partitioning \citep{wegener2003methods}.
The state space \( S \) is divided into \( K+1 \) ranks \((S_0, \ldots, S_K)\) with the following properties: \begin{enumerate}
    \item  
the optimal set $S_{\mathrm{opt}}$ is denoted by $S_0$;
\item the ranking order holds: $f(X_k) > f(X_{k+1})$ for $ X_k \in S_k $ and $ X_{k+1} \in S_{k+1}$ where $0\le k<K$. 
Each rank is referred to as a fitness level. 
\end{enumerate}

The fitness level method uses transition probabilities between fitness levels.
The transition probability from $X_k \in S_k$ to $S_\ell$ is denoted by 
\begin{align*}
p(X_k,S_\ell) =\left\{
\begin{array}{ll}
\ge 0 &\mbox{if } \ell \le k, \\
0 &\mbox{if } \ell > k.
\end{array}
\right.
\end{align*}  
Let $[i,j]$ denote the index set $\{i, i+1, \cdots, j-1,j\}$ and $S_{[i,j]}$ denote the union $ S_i \cup \cdots \cup S_j$
The transition probability from $X_k \in S_k$ to the union $S_{[i,j]}$ is denoted by $p(X_k,S_{[i,j]})$.

On $X^{[t+1]} \in S_{[0,k-1]}$, the conditional transition probability from $X^{[t]}=X_k$ to $X^{[t+1]}\in S_\ell$ (where $\ell <k$) is expressed as  
\begin{align*}
r(X_k,S_\ell) =  \frac{p(X_k, S_\ell)}{p(X_k, S_{[0,k-1]})} =\frac{p(X_k, S_\ell)}{p(X_k, S^+_k)},
\end{align*} 
where $S^+_k$  denotes the union $S_{[0,k-1]}:=S_0 \cup \cdots \cup S_{k-1}$.
 
\subsection{Hitting Time and Hitting Probability} 
Given any two levels $S_k$ and $S_\ell$, the hitting time of a Markov chain refers to the minimum number of generations required for the chain from $X_k \in S_k$ to reach $S_\ell$ \citep{he2025drift,he2025estimation}.

\begin{definition}
Assume that \((X^{[t]})_{t \ge 0}\) be a Markov chain starting from a state \(X_k \in S_k\). The hitting time to reach level \(S_\ell\) is defined as  
$\tau(X_k, S_\ell) = \inf\{t; X^{[t]} \in S_\ell\}.
$  
The mean hitting time \(m(X_k, S_\ell)\) is the expected value of \(\tau(X_k, S_\ell)\).  
The hitting probability from \(X_k\) to \(S_\ell\) is given by  
$
h(X_k, S_\ell) = \Pr(\tau(X_k, S_\ell) < \infty).
$  
\end{definition}

To facilitate analysis,  Table \ref{tab:prob} lists the abbreviations representing minimum and maximum values of transition probabilities, hitting probabilities and mean hitting times used later.  

\begin{table}[htp]
    \centering
    %\resizebox{\columnwidth}{!}{
    \begin{tabular}{c|c }
    \toprule  
          $ p^{{\scriptscriptstyle\min}}_{S_k,S_\ell} :=\displaystyle\min_{X\in S_k}p(X,S_{\ell})$ & 
         $ p^{{\scriptscriptstyle\max}}_{S_k,S_\ell}:=\displaystyle\max_{X\in S_k}p(X,S_{\ell})$  \\\midrule
         $ r^{{\scriptscriptstyle\min}}_{S_k,S_\ell}:=\displaystyle\min_{X\in S_k}r(X,S_{\ell})$ & 
         $ r^{{\scriptscriptstyle\max}}_{S_k,S_\ell}:=\displaystyle\max_{X\in S_k}r(X,S_{\ell})$  \\\midrule
         $ h^{{\scriptscriptstyle\min}}_{S_k,S_\ell}:=\displaystyle\min_{X\in S_{k}}h(X,S_{\ell})$ &
         $ h^{{\scriptscriptstyle\max}}_{S_k,S_\ell}:=\displaystyle\max_{X\in S_{k}}h(X,S_{\ell})$\\\midrule
         $ m^{{\scriptscriptstyle\min}}_{S_k,S_\ell}:=\displaystyle\min_{X\in S_{k}}m(X,S_{\ell})$ &
         $ m^{{\scriptscriptstyle\max}}_{S_k,S_\ell}:=\displaystyle\max_{X\in S_{k}}m(X,S_{\ell})$\\
         \bottomrule
    \end{tabular}%}
    \caption{ Abbreviations for minimum and maximum values}
    \label{tab:prob}
\end{table}

\subsection{Existing Lower Bounds from the Fitness Level Method}  
Existing lower bounds are summarized as follows. \cite{sudholt2012new} assigned a constant coefficient $c_{k,\ell}= c$ (called viscosity) and studied the Type-\( c \) lower bound with the  linear form:
\[
 \frac{1}{p^{\scriptscriptstyle\max}_{S_k, S^+_k}} + \sum_{\ell=1}^{k-1} \frac{c}{p^{\scriptscriptstyle\max}_{S_\ell, S^+_\ell}}, \quad k =1, \cdots, K,
\]
where the constant \( c \) is calculated by 
\begin{equation}
\label{equ:LowerCoeff-c}
c \le \min_{1 < k \le K} \; \min_{1 \le \ell < k} \; \min_{X_k: p(X_k, S_{[0,\ell]}) > 0} \frac{p(X_k, S_{\ell})}{p(X_k, S_{[0,\ell]})}.
\end{equation}
Using a constant $c > 0$, tight lower time bounds are obtained for the (1+1) EA that maximizes various unimodal functions such as LeadingOnes, OneMax, and long $k$-paths \cite{sudholt2012new}.  

\cite{doerr2024lower} assigned coefficients  $c_{k,\ell}=c_{\ell}$  (called visit probability) and  investigated the  Type-\( c_\ell \) lower bound in the linear form:
\[
    \frac{1}{p^{\scriptscriptstyle\max}_{S_k, S^+_k}} + \sum_{\ell=1}^{k-1} \frac{c_{\ell}}{p^{\scriptscriptstyle\max}_{S_\ell, S^+_\ell}}, \quad k =1, \cdots, K,
\] where the coefficient $c_\ell$ can be calculated by
\begin{equation} \label{equ:LowerCoeff-cl}
    c_{\ell} \le \min_{k:\ell < k\le K}  \;   \min_{X_k: p(X_k, S_{[0,\ell]})>0} \frac{p(X_k, S_{\ell})}{p(X_k, S_{[0,\ell]})}. \end{equation} 
Using the coefficient $c_\ell$, tight lower bounds of the (1+1) EA on LeadingOnes,  OneMax, long $k$-paths and jump functions are obtained \cite{doerr2024lower}. 

\cite{he2025drift} proposed drift analysis combined with fitness levels and considered the fitness level method as a special type of drift analysis \citep{he2001drift} and introduced the Type-\( c_{k,\ell} \) lower bound in the following linear form:
\[
   \frac{1}{p^{\scriptscriptstyle\max}_{S_k, S^+_k}} + \sum_{\ell=1}^{k-1} \frac{c_{k,\ell}}{p^{\scriptscriptstyle\max}_{S_\ell, S^+_\ell}}, \quad k =1, \cdots, K,
\]
where the coefficient \( c_{k,\ell} \) is calculated recursively for $k >\ell$,
\begin{align}
\label{equ:LowerCoeff-ckl}
  c_{k,\ell} \le \min_{X_k\in S_k} \left\{ p(X_k, S_\ell) + \sum^{k}_{i=\ell+1}  {p}(X_k, S_i) c_{i,\ell} \right\}.
\end{align}
The above formula is recursive  for calculating coefficients $c_{k,\ell}$ but this recursive calculation is not convenient in applications. The Type-$c$ lower bound is a special case of the formula \eqref{equ:LowerCoeff-ckl} by letting $c_{k,\ell}=c$. The Type-$c_\ell$ bound is a special case of the formula \eqref{equ:LowerCoeff-ckl} by letting $c_{k,\ell}=c_\ell$. However, \cite{he2025drift} points out that the Types-$c$ and $c_\ell$ lower bounds are loose on fitness landscapes with shortcuts. 

\cite{he2025estimation} interpreted each coefficient $c_{k,\ell}$ as a bound on the hitting probability from level $S_k$ to $S_\ell$ and transformed the task of estimating the hitting time into estimating the hitting probability. The formula \eqref{equ:LowerCoeff-ckl} can then be rewritten as a drift condition (see Theorem \ref{the:DriftCondition} in Section \ref{sec:Method} for details). This leads to the drift analysis method for calculating the hitting probability.

%%%%%%%%%%%%%%%%%%%%%%%%%%%%%%%%%%%%%%%%%
\section{Limitations of partitioning the entire search space}
\label{sec:Limitation}
The fitness level method divides the entire search space into several fitness levels. One might expect that exploiting the full search space, rather than a subset of solutions, would leverage additional information and thus improve the lower-bound estimate of the hitting time. This section shows that this expectation does not hold for non‑level‑based fitness functions.
 
\subsection{Level-Based and Non-Level-Based Fitness Functions}
Consider an elitist EA that maximizes a fitness function \( f(x) \) with a fitness level partition \( (S_0, \ldots, S_K) \). We classify fitness functions into two categories.
\begin{enumerate}
  \item A function \( f(x) \) is termed a level-based function for the EA if the maximum and minimum values of transition probabilities between any two fitness levels are equal, i.e.,
\[
\forall \; 0\le \ell <k\le K, \quad  p^{\scriptscriptstyle\max}_{S_k, S_\ell} = p^{\scriptscriptstyle\min}_{S_k, S_\ell}.
\]
The condition is so strong that many functions cannot satisfy it. Only a few artificial fitness functions, such as the OneMax and Jump functions \cite{sudholt2012new}, fall into this category. For level-based fitness functions,  the lower bound \eqref{equ:Lower-bound} and  upper bound \eqref{equ:Upper-bound} are equal which represent the acuate mean hitting time.
  
  \item A function \( f(x) \) is termed a non-level-based function if the maximum and minimum values of transition probabilities between some fitness levels differ, i.e.,
\[
\exists\; 0\le \ell <k<K, \quad  p^{\scriptscriptstyle\max}_{S_k, S_\ell} > p^{\scriptscriptstyle\min}_{S_k, S_\ell}.
\]
In real-world optimization problems, many fitness functions do not follow a level-based structure, such as the knapsack problem instances P1 to P6 in the next section. For non‑level‑based fitness functions, the lower bound \eqref{equ:Lower-bound} and the upper bound \eqref{equ:Upper-bound} differ, implying that at least one of them is loose. Interestingly, we find that the lower bound \eqref{equ:Lower-bound} usually is loose for the hitting time, whereas the upper bound \eqref{equ:Upper-bound} is tight.

\end{enumerate}

%%%%%%%%%%%%%%%%%%%%%%%%%%%%%%%%%%%%%%%%%%%%%%%
\subsection{Background Optimization Problem}
 In this paper, we use the knapsack problem as the background optimization problem because it a classic benchmark problem in evolutionary computation \citep{michalewicz1996genetic}. The problem involves \( n \) items where each item $x_i$ is associated with a specific weight \( w_i \) and value \( v_i \). The objective is to select a subset of these items to  be placed in the knapsack, ensuring that the total weight does not exceed the capacity \( C \) while maximizing the total value. For the \( i \)th item, \( b_i = 1 \) indicates the item is included in the knapsack, whereas \( b_i = 0 \) indicates it is not. Mathematically, the knapsack problem can be expressed as a constrained optimization problem: let $x=(x_1, \ldots, x_n)$,
\begin{align}
\label{equ:Knapsack}
\begin{array}{ll}
&\max f(x) = \sum^n_{i=1} v_i b_i, \quad \text{subject to } \sum^n_{i=1} w_i b_i \leq C. 
\end{array}
\end{align}

\begin{algorithm}[ht]
\caption{The (1+1) EA using the Death Penalty Method}
\label{alg2}
\begin{algorithmic}[1] 
\STATE   let the initial solution $ X^{[0]} =x$ be an empty knapsack;
\FOR{$t=1,2,\cdots$}
\STATE   flip  each bit of $X^{[t-1]}$ independently with probability $1/n$ and generate a new solution $y$;
\IF{$y$ is feasible and $f(y)\ge f(x)$}
\STATE let $X^{[t]}= y$;
\ELSE
\STATE let $X^{[t]}=x$.
\ENDIF
\ENDFOR
\end{algorithmic}
\end{algorithm} 

To simplify the analysis, we consider a (1+1) EA that uses the death penalty method to deal with the constraint. Any infeasible solution is discarded.  Algorithm~\ref{alg2} details its procedural implementation.  The (1+1) EA combines bitwise mutation and elitist selection.  It is widely used in theoretical research on EAs \citep{droste2002analysis,sudholt2012new,friedrich2023analysis,doerr2024lower}.

\begin{table*}[ht]
    \centering\resizebox{\columnwidth}{!}{
    \begin{tabular}{ccccc cccc}
    \toprule
          &  item $i$ &  $1$ & $2$ &$3,\ldots, n$ & capacity $C$ &global optimum &local optimum  \\
            \midrule
             \multirow{2}{*}{P1} &  value $v_i$   & $n-2$ & $n/2$ & $1$ &   \multirow{2}{*}{$n-2$} & \multirow{2}{*}{$L_{(1, 0;0)}$ and $L_{(0,0;n-2)}$} & \multirow{2}{*}{$L_{(0,1;1)}$} \\ 
           & weight $w_i$ & $n-2$ & $n-3$ &$1$ &   \\ 
                       \midrule
             \multirow{2}{*}{P2} &  value $v_i$   & $n-2$ & $n/2$ & $1$ &   \multirow{2}{*}{$n-3$} & \multirow{2}{*}{ $L_{(0,0;n-3)}$} & \multirow{2}{*}{$L_{(0,1;0)}$} \\ 
            & weight $w_i$ & $n-2$ & $n-3$ & $1$ &   \\ 
                              \midrule
            \multirow{2}{*}{P3}  & value $v_i$   & $n-2 $ & $n/2 $ & $1$ &   \multirow{2}{*}{$n-2$} & \multirow{2}{*}{$L_{(1, 0;1)}$  } & \multirow{2}{*}{$L_{(0,1;1)}$ and $L_{(0,0;n-2)}$}
            \\ 
           & weight $w_i$ & $n-3$ & $n-3$ & $1$ &   
            \\ 
            \midrule
                         \multirow{2}{*}{P4} &  value $v_i$   & $n-2$ & $n/2$ & $1$ &   \multirow{2}{*}{$n-2$} & \multirow{2}{*}{$L_{(0,1;3n/4-2)}$}  & \multirow{2}{*}{$L_{(1, 0;0)} \cup L_{(0,0;n-2)}$}  \\ 
           & weight $w_i$ & $n-2$ & $n/4$ &$1$ &   \\    \midrule
             \multirow{2}{*}{P5} &  value $v_i$   & $n-2$ & $n/2-1$ & $1$ &   \multirow{2}{*}{$n-2$} & \multirow{2}{*}{$L_{(0,0;n-2)}$}  & \multirow{2}{*}{ $L_{(0,1;n/2-2)}$}  \\ 
           & weight $w_i$ & $n-1$ & $n/2$ &$1$ &   \\  
           \midrule
             \multirow{2}{*}{P6} &  value $v_i$   & $n-2$ & $n/2$ & $1$ &   \multirow{2}{*}{$+\infty$} & \multirow{2}{*}{ $L_{(1,1;n-2)}$} & \multirow{2}{*}{ --} \\ 
           & weight $w_i$ & $n-2$ & $n-3$ &$1$ &   \\ 
            \bottomrule
    \end{tabular}
    }
    \caption{Knapsack problem instances.}
    \label{tab:knapsack}
\end{table*} 

Six knapsack problem instances are presented in Table \ref{tab:knapsack} as a benchmark suite. Small changes in value, weight, or capacity will change the position of global and local optima.  The following notations are used in the analysis of these knapsack instances. 
\begin{align*}
    &L_{(b_1, b_2; i)} := \{x = (b_1, b_2, \ldots, b_n); b_3 + \cdots + b_n=i\}.\\ 
        &L^+_{(b_1, b_2;i)} := \{x ;  f(x) >f(b_1, b_2;i) \}.\\
    &L_{(b_1, b_2; [i_1, i_2])}:= \{x=  (b_1, b_2, \ldots, b_n); i_1 \leq b_3 + \cdots + b_n \leq i_2\}.\\
    &p_{(a_1,a_2;i),(b_1,b_2;j)}:=p(x \in L_{(a_1,a_2;i)}, L_{(b_1,b_2;j)}).\\ 
        &p_{(b_1, b_2;i),(b_1, b_2;i)^+}:=p(x \in L_{(b_1, b_2;i)}, L^+_{(b_1, b_2;i)}).\\
    &h_{(a_1,a_2;i),(b_1,b_2;j)}:=h(x \in L_{(a_1,a_2;i)}, L_{(b_1,b_2;j)}).\\ 
    &m_{(a_1,a_2;i),(b_1,b_2;j)}:=m(x \in L_{(a_1,a_2;i)}, L_{(b_1,b_2;j)}).\\
 &p^{{\scriptscriptstyle\min}}_{(a_1,a_2;[i_1,i_2]),(b_1,b_2;[j_1,j_2])}:= \displaystyle \min_{x \in L_{(a_1,a_2;[i_1,i_2])}}p(x , L_{(b_1,b_2;[j_1,j_2])}).\\ &p^{{\scriptscriptstyle\max}}_{(a_1,a_2;[i_1,i_2]),(b_1,b_2;[j_1,j_2])}:= \displaystyle \max_{x \in L_{(a_1,a_2;[i_1,i_2])}}p(x, L_{(b_1,b_2;[j_1,j_2])}). 
\end{align*}

For example, the notation \( L_{(1, 0; i)} \) denotes the set of knapsacks where $b_1=1,$ $ b_2=0$ and $b_3+\cdots+b_n=i$. The cardinality of this set is $\binom{n-2}{i}$. The transition probability from \( x \in L_{(1, 0; i)} \)  to set \( L_{(1, 0; n-2)} \) is denoted by $p_{(1, 0; i), (1, 0; n-2)}$ in short.

%\begin{table}[htb]
%    \centering
%    %\resizebox{\columnwidth}{!}{
%    \begin{tabular}{l}
%    \toprule 
%    $L_{(b_1, b_2; i)} := \{x = (b_1, b_2, \ldots, b_n); b_3 + \cdots + b_n=i\}$\\
%    \midrule
%    $ L_{(b_1, b_2; [i_1, i_2])}:= \{x=  (b_1, b_2, \ldots, b_n); i_1 \leq b_3 + \cdots + b_n \leq i_2\}$\\\midrule
 %$p_{(a_1,a_2;i),(b_1,b_2;j)}:=p(x \in L_{(a_1,a_2;i)}, L_{(b_1,b_2;j)})$\\
 %\midrule    $h_{(a_1,a_2;i),(b_1,b_2;j)}:=h(x \in L_{(a_1,a_2;i)}, L_{(b_1,b_2;j)})$\\ 
%\midrule    $m_{(a_1,a_2;i),(b_1,b_2;j)}:=m(x \in L_{(a_1,a_2;i)}, L_{(b_1,b_2;j)})$\\ 
% \midrule
% $p^{{\scriptscriptstyle\min}}_{(a_1,a_2;[i_1,i_2]),(b_1,b_2;[j_1,j_2])}:= \displaystyle \min_{x \in L_{(a_1,a_2;[i_1,i_2])}}p(x , L_{(b_1,b_2;[j_1,j_2])})$\\\midrule   %$p^{{\scriptscriptstyle\max}}_{(a_1,a_2;[i_1,i_2]),(b_1,b_2;[j_1,j_2])}:=\displaystyle \max_{X \in L_{(a_1,a_2;[i_1,i_2])}}p(X, L_{(b_1,b_2;[j_1,j_2])})$\\\midrule
%$p^{{\scriptscriptstyle\max}}_{(a_1,a_2;[i_1,i_2]),(b_1,b_2;[j_1,j_2])}:= \displaystyle \max_{x \in L_{(a_1,a_2;[i_1,i_2])}}p(x, L_{(b_1,b_2;[j_1,j_2])})$\\
%         \bottomrule
%    \end{tabular}%} 
%\caption{Notations in knapsack instances.}
%    \label{tab:Notations}
%\end{table} 

\subsection{Visualization via Directed Graphs}
\begin{figure*}[ht]
    \centering
\includegraphics[width=0.48\textwidth]{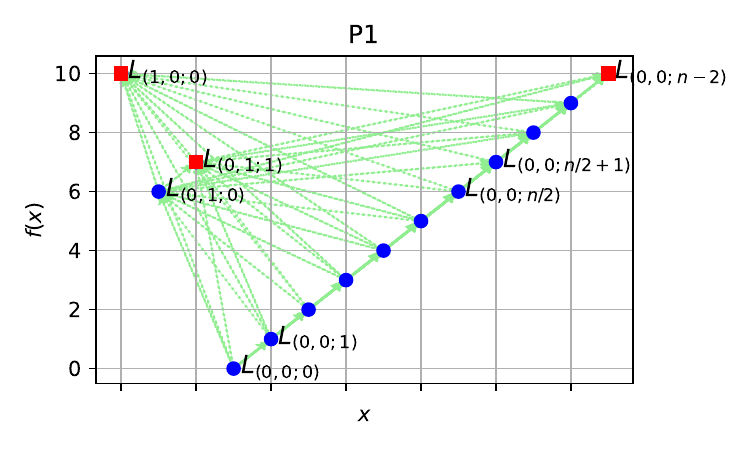}
\includegraphics[width=0.48\textwidth]{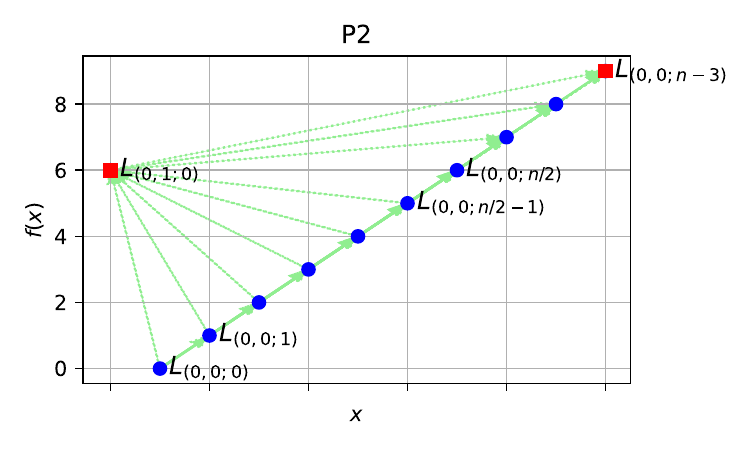} \\  \includegraphics[width=0.48\textwidth]{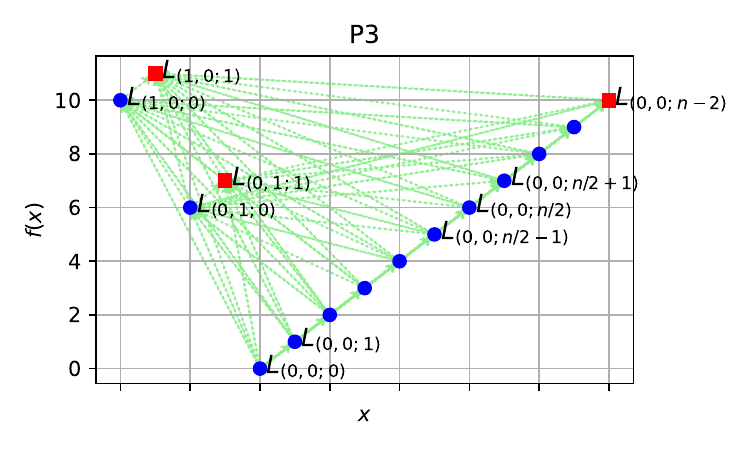}
\includegraphics[width=0.48\textwidth]{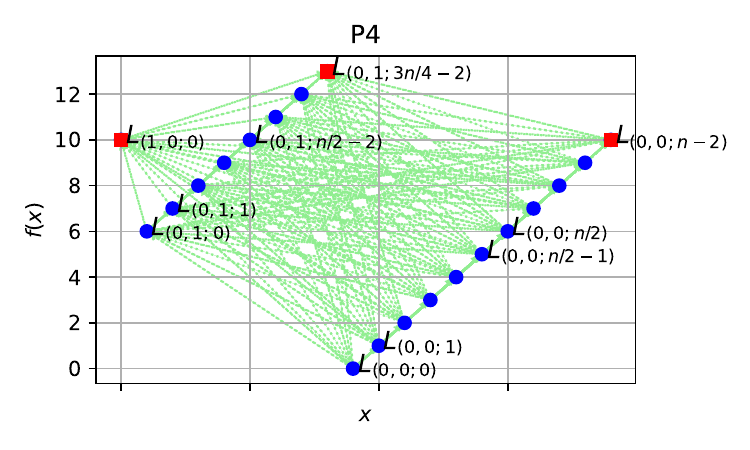}  \\  \includegraphics[width=0.48\textwidth]{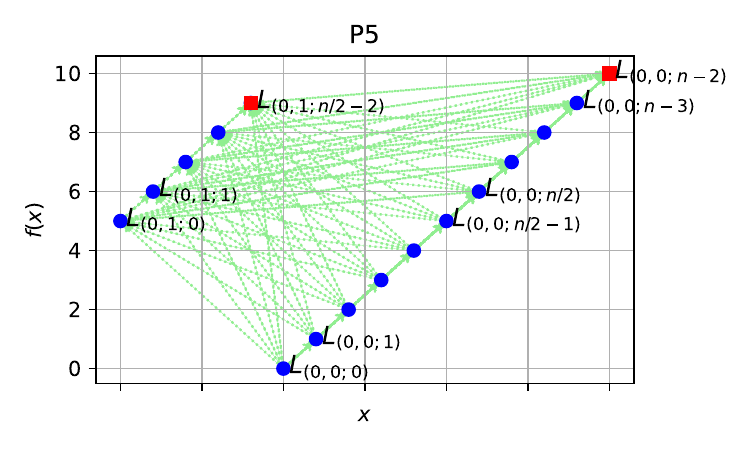}
\includegraphics[width=0.48\textwidth]{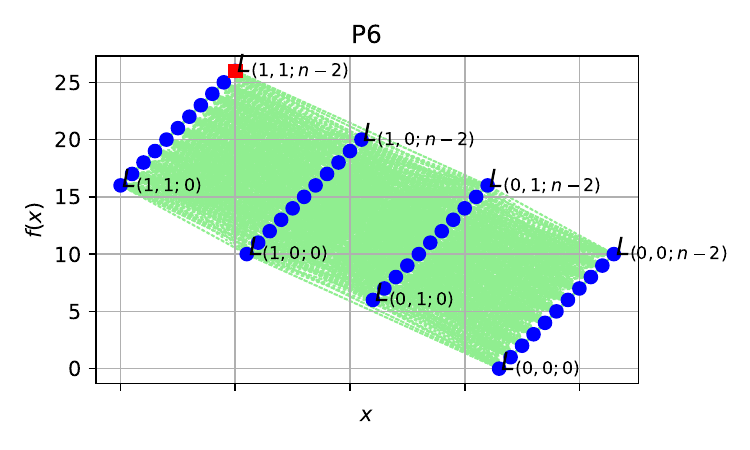} 
        \caption{Directed graph of (1+1) EAs on instances P1 to P6. Vertices represent  states (feasible solutions). Arcs represent transitions. $n=12$.}
    \label{fig:Knapsack}
\end{figure*}

The dynamics of an  elitist EA can be visualized as a Markov chain over a directed graph  \( G = (V, A) \) \citep{chong2019coevolutionary,chong2019new}. Each vertex \( k \in V \) corresponds to a level  \( S_k \) for \( k = 0, \ldots, K \). The set of arcs \( A \) corresponds to transitions, where an arc \( (k, \ell) \in A \)  means that the conditional transition probability $r^{\scriptscriptstyle\max}_{ S_k,S_\ell} >0$. To visually demonstrate detailed transitions, a level is sometimes depicted as multiple  vertices located at different positions; however, these vertices are analytically treated as a single vertex, and no arcs are drawn between these vertices at the same fitness level. For example, in Figure~\ref{fig:Knapsack}.P2, two vertices $L_{(0,1;0)}$ and $L_{(0,0;n/2)}$ belong to the same fitness level and are considered as one vertex in the fitness level method. 

A {path} from $S_k$ to $S_\ell$, denoted by $P[\ell,k]$, is an acyclic sequence of vertices $\{k_0, k_1, \cdots, k_m\}$ such that $k_0=\ell,$ $k_m=k$ and every pair $(k_{i},k_{i-1})$  is an arc. This path can be explicitly written as $k_m \to k_{m-1} \to \cdots \to k_1\to k_0$.  For example, in Figure~\ref{fig:Knapsack}.P2, a path is $L_{(0,1;0)} \to \cdots L_{(0,0;n/2)} \to \cdots L_{(0,0;n-3)}$.    The notation \(P(\ell, k] \) represents the subpath of \(P[\ell, k] \) after deleting the vertex \(\ell \). The notation \(P[\ell, k)\) represents the subpath of \(P[\ell, k] \) after deleting the vertex \(k \).  
The complete path $k\to k-1 \to \cdots \to \ell+1 \to \ell$ is includes all level from $k$ to $\ell$. It is denoted as $[\ell,k]$, the same as the index set.
  
The notation \( P_{\mathrm{sub}}[\ell, k] \) denotes a subsequence of path \( P[\ell, k] \) consisting of endpoints \( \ell \) and \( k \), and a subset of intermediate vertices from \( P[\ell, k] \), for example, $P_{\mathrm{sub}}[\ell, k]= k \to \ell+1 \to \ell$. This subsequence relation is denoted by \( P_{\mathrm{sub}}[\ell, k] \subseteq P[\ell, k] \). Note:  a subsequence \( P_{\mathrm{sub}}[\ell, k] \) may not be a path from $S_k$ to $S_\ell$.  

A path $P[\ell,k]$ can be decomposed into multiple segments, e.g., two segments $ P[\lambda, k]= k  \to  \cdots \to  \lambda$ and $P[\ell, \lambda]= \lambda \to \cdots \to \ell$.

Figure~\ref{fig:Knapsack} illustrates the directed graphs of the \((1+1)\) EA for instances P1 to P6.  Note that this drawing does not represent the actual Hamming distance between the two states. For instance P1, the Hamming distance between the local optimum $L_{(0,1;1)}$ and a global optimum $L_{(1,0;0)}$ is 3. There is no local optimum in P6.  P1 and P6 are easy for the (1+1) EA. But in each problem instance from P2 to P5, there exists a wide gap between a local optimum and the global optimal set, and the Hamming distance between them is \(\Omega(n)\). The (1+1) EA performs poorly on these instances because the wide gap requires a jump that is unlikely to occur.

All instances belong to non-level-based fitness functions. Let us demonstrate this by instance P1. A similar analysis can be applied to  each problem instance from P2 to P6. 
Figure~\ref{fig:Knapsack}.P1 shows that for P1, the fitness level \( f(x) = n/2+1 \) consists of  \( L_{(0,1;1)} \) and \( L_{(0,0;n/2+1)} \). The higher level  \( f(x) > n/2+1 \) consists of \( L_{(0,0;[0, n/2+2])} \cup L_{(1,0;0)}\). 

On the one hand, the transition from \( L_{(0,1;1)} \) to \( L_{(0,0;[0, n/2+2])} \cup L_{(1,0;0)}\) occurs only if (i) bit \( b_2 \) is flipped and \( n/2+1 \) of the \( n-3 \) zero-valued bits in \([b_3, b_n]\) are flipped or (ii) bits $b_1, b_2$ and the unique one-valued bit in $[b_3,b_n]$ are flipped. This transition has a probability 
\begin{align}
\label{eq:P1-min}
p_{(0,1;1), (0,0;[0, n/2+2]) \cup (1,0;0)} & \le \frac{1}{n} \binom{n-3}{ {n}/{2}+1} \left(\frac{1}{n}\right)^{{n}/{2}+1} + \left(\frac{1}{n}\right)^3 \le  O\left(n^{-3}\right). 
\end{align}

 On the other hand, the  transition from \( L_{(0,0; n/2+1)} \) to \( L_{(0,0;[0, n/2+2])} \cup L_{(1,0;0)}\) occurs if (i) one of the \( n/2-3 \) zero-valued bits in \([b_3, b_n]\) is flipped while the other bits remain unchanged, or (ii) bits $b_1$ and all $n/2+1$ one-valued bits in $[b_3,b_n]$ are flipped while the other bits remain unchanged. This transition has a probability 
\begin{align}
p_{(0,0; {n}/{2}+1),(0,0;[0, {n}/{2}+2]) \cup (1,0;0)} &\ge  \frac{{n}/{2}-3}{n} \left(1-\frac{1}{n}\right)^{n-1} + \left(\frac{1}{n}\right)^{{n}/{2}+2} \left(1-\frac{1}{n}\right)^{\frac{n}{2}-2}\nonumber\\
&= \Omega(1). 
\label{eq:P1-max}
\end{align}

Comparing \eqref{eq:P1-max} with \eqref{eq:P1-min}, we find that the maximum value of the transition probability from the fitness level \( f(x) = n/2+1 \) to the higher level \( f(x) > n/2+1 \) is larger than the minimum value. Thus, the fitness function of P1 is not level-based.

%%%%%%%%%%%%%%%%%%%%%%%%%%%%%
\subsection{Loose Lower bounds on Non-level-based Fitness Functions} 
Taking P1 as an example, we prove that the lower bound \eqref{equ:Lower-bound} is up to $O(n\log n)$.
We divide the search space \(S\) into the following fitness levels. Each level represents a fitness value.
\[
S_{\ell} = \left\{
\begin{array}{lll}
L_{(1,0;0)} \cup L_{(0,0;n-2)}, & \ell = 0, \\
L_{(0,0;n-2-\ell)}, & \ell = 1, \cdots, {n}/{2}-1, \\
L_{(0,1;1)} \cup L_{(0,0;n-2-\ell)}, & \ell = {n}/{2}-3,\\
L_{(0,1;0)} \cup L_{(0,0;n-2-\ell)}, & \ell = {n}/{2}-2,\\
L_{(0,0;n-2-\ell)}, & \ell = n-2.
\end{array}
\right.
\] 

The lower bound \eqref{equ:Lower-bound} on the mean hitting time from the empty knapsack $S_{n-2}$ to the optimal solutions $S_0$ is up to
\begin{align}
 & \frac{1}{p^{\scriptscriptstyle\max}_{S_{n-2}, S_{[0,n-3]}}} + \sum_{\ell=1}^{n-3} \frac{h^{\min}_{S_{n-2},S_\ell}}{p^{\scriptscriptstyle\max}_{S_\ell, S_{[0,\ell-1]}}} \le  \frac{1}{p^{\scriptscriptstyle\max}_{S_{n-2}, S_{[0,n-3]}}} + \sum_{\ell=1}^{n-3} \frac{1}{p^{\scriptscriptstyle\max}_{S_\ell, S_{[0,\ell-1]}}}. \label{eq:P1-1}
\end{align}

For any $\ell >1$, the level $S_\ell$ includes the state $L_{(0,0;n-2-\ell)}$. 
The transition from $L_{(0,0;n-2-\ell)}$ to  $L_{(0,0;[0,n-2-\ell))}$ happens only if one of $\ell$ zero-valued bits in $[b_3,b_n]$ is flipped and other bits are unchanged.  
\begin{align} 
p^{\scriptscriptstyle\max}_{S_\ell, S_{[0,\ell-1]}} \ge  p_{(0,0;n-2-\ell),(0,0;[0,n-2-\ell))} \ge    \frac{\ell}{n} \left(1-\frac{1}{n}\right)^{n-1} \ge \frac{\ell}{ne}.
\label{eq:P1-2}
\end{align} 

By inserting  \eqref{eq:P1-2} to \eqref{eq:P1-1}, we get
\begin{align}  \label{ieq:P1-3}
 \frac{1}{p^{\scriptscriptstyle\max}_{S_{n-2}, S_{[0,n-3]}}} + \sum_{\ell=1}^{n-3} \frac{h^{\min}_{S_{n-2},S_\ell}}{p^{\scriptscriptstyle\max}_{S_\ell, S_{[0,\ell-1]}}}  \le  1 + \sum_{\ell=1}^{n-3} \frac{ne}{\ell}  =O(n \log n). 
\end{align}   
This means that, the lower bound \eqref{equ:Lower-bound} is no more than $O(n \log n)$. For instances P2 to P6, we can perform a similar analysis as P1 and come to the same conclusion: using the fitness level method, we cannot get a lower bound better than $O(n \log n)$.  We omit the detailed proof. The lower bound $O(n \log n)$ is trivial, since it is well-known that  the mean hitting time of the (1+1) EA is $\Omega(n \log n)$ on all fitness functions with one optimum \citep{sudholt2012new,he2015easiest}. 
%The problem arises because for most non-level-based fitness functions, there is a significant difference between the maximum and minimum transition probabilities across some fitness levels.  

%%%%%%%%%%%%%%%%%%%%%%%%%
%%%%%%%%%%%%%%%%%%%%%%%%%

 %%%%%%%%%%%%%%%%%%%%%%%%%%%%%%

%%%%%%%%%%%%%%%%%%%%%%%%%%%%%%%%%%%
\section{The Subset Fitness Level Method}
\label{sec:Method}
To overcome the limitations of partitioning the entire search space, this section proposes a new subset fitness level method which partitions a subset of non-optimal solutions into fitness levels. This method is designed to produce a tight lower bound on the hitting time.

%%%%%%%%%%%%%%%%%%%%%%%%%%%%%%%%%%%%
\subsection{Level Partitioning} 
The subset method is based on the partitioning of a non-optimal subset into fitness levels instead of the partitioning of the entire search space into  fitness levels.

\begin{definition}
Let \( S' \) be a subset of non-optimal states, partitioned into \( K \) fitness levels \( (S_1, \ldots, S_K) \) according to decreasing fitness values. Define \( S_0 = S \setminus S' \), which contains the remaining states but is not considered a fitness level. The collection \( (S_0, S_1, \ldots, S_K) \) forms a \emph{level partition} of the state space, where each subset $S_\ell$ is called a \emph{level}. 
\end{definition} 
 
Under the above level partitioning, level \(S_0\) may contain some non-optimal solutions that are worse than the solutions in other levels. The choice of the non-optimal subset \(S'\) is flexible.  For a multimodal function, it is beneficial to include local optima in the subset \(S'\) since elitist EAs often spend a significant amount of time in local optima. This long-term stagnation near local optima is  a major contributor to the mean hitting time. The subset \(S'\) usually contain one or more paths from the initial state to the local optimum.  If the subset \(S'\) is selected as the set of all non-optimal states, it degenerates into the traditional fitness level method. 

\begin{figure}[ht]
    \centering
    \includegraphics[width=0.48\textwidth]{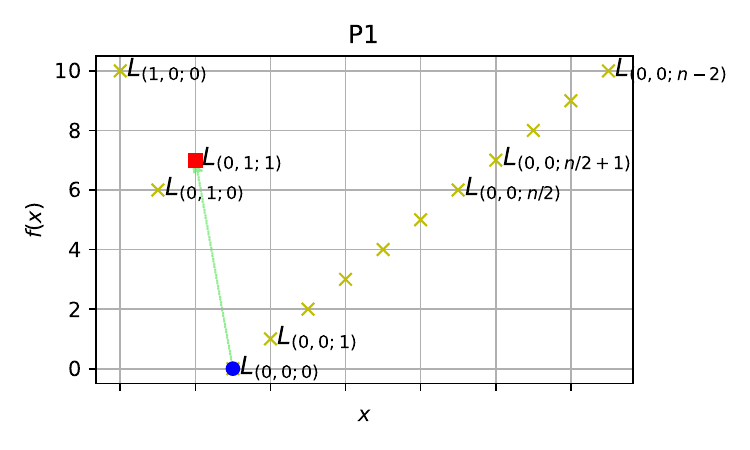}
\includegraphics[width=0.48\textwidth]{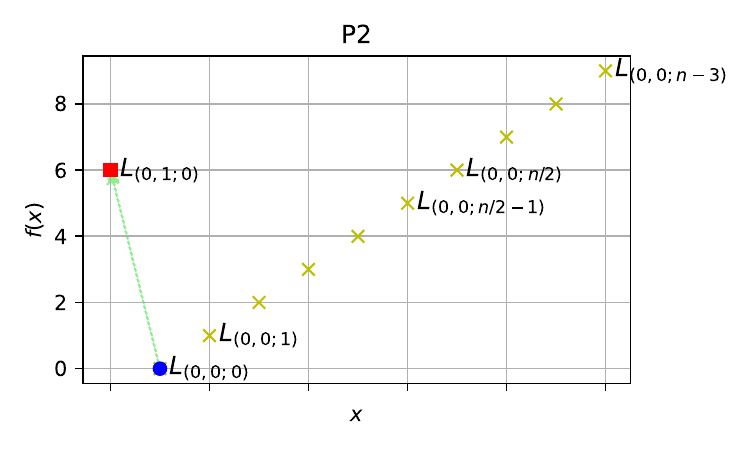}\\
\includegraphics[width=0.48\textwidth]{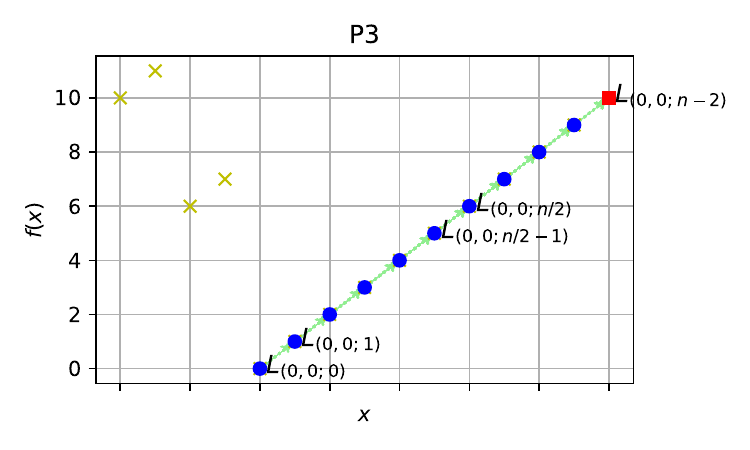}
\includegraphics[width=0.48\textwidth]{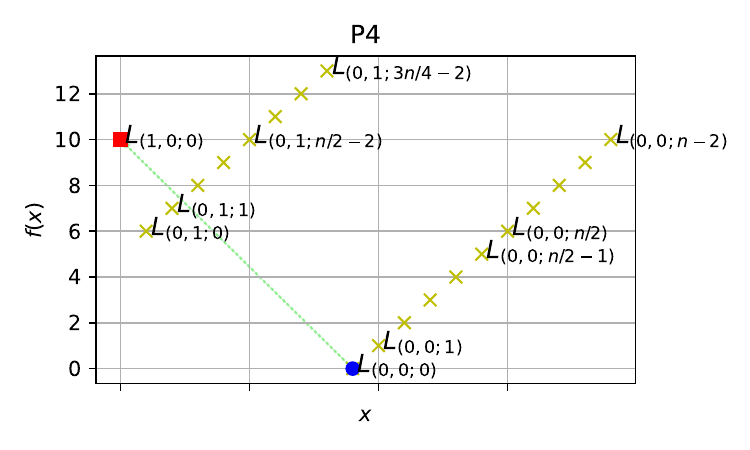}\\
\includegraphics[width=0.48\textwidth]{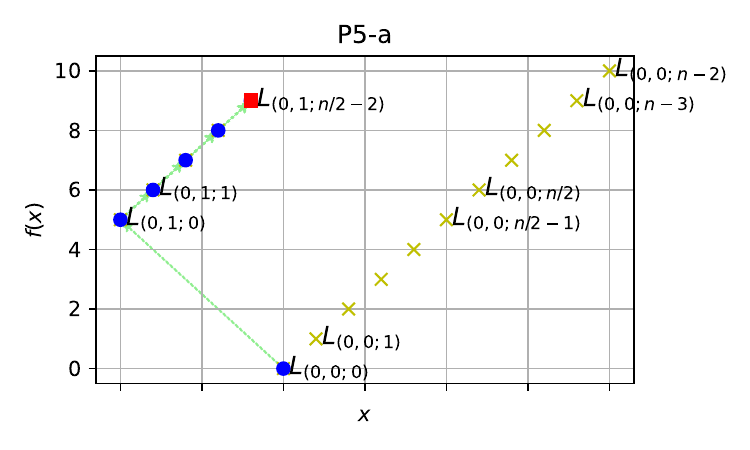}
\includegraphics[width=0.48\textwidth]{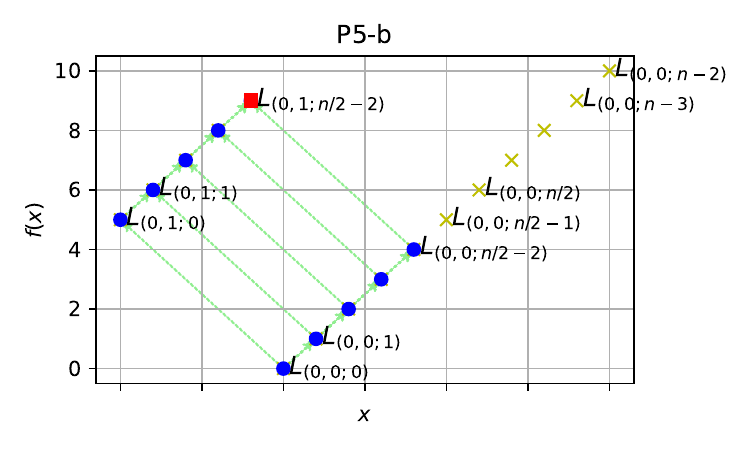}
        \caption{Selected subsets and paths in Instances P1 to P5. Vertices represent  states. Arcs represent transitions. $n=12$. }
    \label{fig:Subset}
\end{figure}

Figure~\ref{fig:Subset} illustrates the subsets for instances P1 through P5 that are used in the analysis presented in the next section. In particular, Figure~\ref{fig:Subset}.P5‑a shows a single path from the empty‑knapsack state \(L_{(0,0;0)}\) to the local optimum \(L_{(0,1;n/2-2)}\), whereas Figure~\ref{fig:Subset}.P5‑b displays multiple paths leading to the same local optimum.

An auxiliary absorbing Markov chain is constructed whose transition probabilities (denoted by $p'(X,Y)$) satisfy
\begin{equation}
\label{equ:Markov-chain2}
   p'(X,Y) =
    \left\{\begin{array}{ll}
         p(X,Y) & \mbox{if } X \in S', \\
          0 & \mbox{if } X  \in S \setminus S', Y \in S. 
    \end{array}
    \right.
\end{equation} 
In the above chain, $S_0$ is regarded as an absorbing state. 
The lemma below states that the mean hitting time  \( m(X_k, S_0) \) is a lower bound on the mean hitting time to the optimal solution set \( m(X_k, S_{\mathrm{opt}}) \).

\begin{lemma} 
\label{lemma:subgraph}
Given a level partition $(S_0, S_1, \ldots, S_K)$, the mean hitting time of the EA from an $X$ to  $S_{\mathrm{opt}}$ is not less tan than the hitting time from $X$ to $S_0$.
\end{lemma} 

\begin{proof}
Since $S_0 \supset S_{\mathrm{opt}}$, the event of arriving at $S_{\mathrm{opt}}$ implies the event of reaching $S_0$ and then the conclusion is derived from the definition of hitting time.
\end{proof}

%%%%%%%%%%%%%%%%%%%%%%%%%%

\subsection{Drift Analysis of Hitting Probability}
\label{secDrift}  
Recently,  drift analysis with fitness levels was proposed to estimate the hitting time of elitist EAs \citep{he2025drift,he2025estimation}. Although the method was originally formulated using fitness level partitioning, it can be naturally extended to level partitioning. Since the level partitioning is based on a subset of the search space, we call this new adaptation the subset fitness level method.  The following theorem shows that the task of estimating the mean hitting time can be transformed into a task of estimating the hitting probability.  It is a subset adaption of Theorem 2 in \citep{he2025estimation}.

\begin{theorem}\label{the:HittingTime}
Given an elitist EA and a level partition $(S_0, \ldots, S_K)$, the mean hitting time from $S_k$ to $S_0$ satisfies 
        \begin{align}
          &  m^{{\scriptscriptstyle\min}}_{S_k,S_0}\ge  \frac{1}{p^{{\scriptscriptstyle\max}}_{S_k,S^+_{k}}}+\sum_{\ell =1}^{k-1}\frac{  h^{{\scriptscriptstyle\min}}_{S_k,S_\ell} }{ p^{{\scriptscriptstyle\max}}_{S_\ell,S^+_{ \ell  }}}\label{ieq:lb-ht}.
        \end{align}  
\end{theorem} In most cases, it is difficult to calculate the accurate value of the hitting probability $h_{S_k,S_\ell} $. Instead, we aim at estimating a lower bound coefficient $c_{k,\ell}\le h^{{\scriptscriptstyle\min}}_{S_k,S_\ell}$ by drift analysis. A drift function \(c(X_k, S_\ell) \) is used to approximate the hitting probability \(h(X_k, S_\ell) \). 

\begin{definition}
Given an elitist EA, a level partition $(S_0, \ldots, S_K)$, for $\ell, k =1, \ldots, K$, the \emph{drift function} is defined as for any $x_k \in S_k$,
\begin{equation}
\label{equ:DriftFunction}
c(X_k, S_\ell) := c_{k,\ell} =
\begin{cases}
0 & \text{if } \ell > k,\\
1 & \text{if } \ell = k,\\
\in [0,1] & \text{if } \ell < k,
\end{cases}
\end{equation}
where $c_{k,\ell}$ is the linear bound coefficient. 
The corresponding \emph{conditional drift} from $X_k$ to $S_\ell$  is defined as
\begin{align}
\tilde{\Delta} c(X_k, S_\ell):= c_{k,\ell}-\sum^{K}_{i=0} \sum_{Y_i \in S_i}  r(X_k, Y_i)   c_{i, \ell} = c_{k,\ell}- \sum^{k-1}_{i=\ell+1}  {r}(X_k, S_i) c_{i,\ell} \label{equ:ConditionalDrift},
\end{align}where we exploit the fact that $r(X_k, S_i)=0$ for $i \ge k$, $c_{\ell,\ell}=1$ and   $c_{i, \ell}=0$ for $i< \ell$.
\end{definition}

The following theorem establishes the drift condition to determine a lower bound on the hitting probability. It is a subset adaption of Theorem 3 in \citep{he2025estimation}.  
\begin{theorem}  
\label{the:DriftCondition}   Given an elitist EA and a level partition $(S_0, \ldots, S_K)$, for two levels $S_\ell, S_k: 1 \le \ell < k \le K$,   if a drift function \eqref{equ:DriftFunction} satisfies that for any $ \ell <j \le k$ and any $X_j \in S_j$,   the conditional drift $\tilde{\Delta} c(X_j, S_\ell)\le 0$, 
then   the hitting probability $h^{{\scriptscriptstyle\min}}_{S_k,S_\ell} \ge c_{k,\ell}$.  
\end{theorem}

%The above drift analysis theorem is used to estimate the hitting probability, not the hitting time.
The condition that for all \(  X_k \in S_k:  \tilde{\Delta} c(X_k, S_\ell) \le 0 \) %the two drift conditions 
can be restated as 
\begin{align} 
 c_{k,\ell} \le \min_{X_k \in S_k}  \sum^{k-1}_{i=\ell}  {r}(X_k, S_i) c_{i,\ell} .
\label{equ:DriftCondition2}
\end{align} So we can directly obtain the following  formula to calculate the coefficient $c_{k, \ell}$ recursively from coefficients $c_{\ell,\ell}, \cdots, c_{k-1, \ell}$.

%%%%%%%%%%%%%%%% 
\begin{corollary}
\label{cor:recursive}
    Given an elitist EA and a level partition $(S_0, \ldots, S_K)$, for two levels $S_\ell, S_k: 1 \le \ell < k \le K$,  if   for $ \ell <j \le k$, coefficient
    \begin{align}
    \label{equ:recursive}
        c_{j,\ell} \le  \sum^{j-1}_{i=\ell}  r^{\scriptscriptstyle\min}_{S_j, S_i} c_{i,\ell},
    \end{align}
then the hitting probability $h^{{\scriptscriptstyle\min}}_{S_k,S_\ell} \ge c_{k,\ell}$.  
\end{corollary}

According to the above corollary, we get the best  coefficients  as $ c^{\scriptscriptstyle\min}_{\ell, \ell}= 1$ and 
\begin{align} \label{eq:recursion}
 c^{\scriptscriptstyle\min}_{j, \ell} = \sum^{j-1}_{i=\ell}  r^{\scriptscriptstyle\min}_{S_j, S_i} c_{i,\ell}, \quad  j=\ell+1, \ldots, k. 
\end{align}  
By Corollary \ref{cor:recursive}, any coefficient \( c_{j, \ell} \) satisfying \eqref{equ:recursive} is bounded above by \( c^{\scriptscriptstyle\min}_{j, \ell} \). Therefore, \(c^{\scriptscriptstyle\min}_{j, \ell} \) represents the best coefficient among all those obtained by drift analysis of hitting probability. Note: we always have \( h^{\min}_{S_j,_\ell} \ge c^{\min}_{j, \ell} \), but we cannot conclude that \( h^{\min}_{S_j,S_\ell} = c^{\min}_{j, \ell} \).

\subsection{Explicit Formulas} 
To avoid recursive computation in \eqref{eq:recursion}, it is preferable to derive explicit formulas for the lower bound coefficients.  
By inductive reasoning on the recursive relation in \eqref{eq:recursion}, we obtain the following explicit expression:
\begin{align}
\label{eq:expression-sum2}
c^{\scriptscriptstyle\min}_{j,\ell}
= r^{\scriptscriptstyle\min}_{S_j, S_\ell} + \sum_{\ell < j_1 < j} r^{\scriptscriptstyle\min}_{S_j, S_{j_1}} r^{\scriptscriptstyle\min}_{S_{j_1}, S_\ell} + \sum_{\ell < j_1 < j_2 < j} r^{\scriptscriptstyle\min}_{S_j, S_{j_2}} r^{\scriptscriptstyle\min}_{S_{j_2}, S_{j_1}} r^{\scriptscriptstyle\min}_{S_{j_1}, S_\ell} + \cdots
\end{align}
Each term in this expansion corresponds to a product of conditional transition probabilities along a subsequence from \( S_j \) to \( S_\ell \). The best coefficient \( c^{\scriptscriptstyle\min}_{j,\ell} \)  can be interpreted as the cumulative conditional transition probability of all subsequences from \(S_j \) to \(S_\ell \).  We reformulate the formula \eqref{eq:expression-sum2} using subsequence notation, leading to the following theorem.

\begin{theorem}\label{the:ExplicitExpression0}
Given an elitist EA, a level partition $(S_0, \ldots, S_K)$, and two levels $S_\ell,  S_k: 1 \le \ell < k \le K$,  for any $\ell <j \le k$, let the path $[\ell,j]=\{\ell, \ell+1, \cdots, j-1, j\}$.  Define the best coefficients as   $c^{\scriptscriptstyle\min}_{\ell,\ell}=1$ and  
\begin{align} \label{eq:expression-sum}
 c^{\scriptscriptstyle\min}_{j, \ell}=  
 \sum_{P_{\mathrm{sub}}[\ell,j] \subseteq [\ell,j]}  \, r^{\scriptscriptstyle\min}_{S_{j}, S_{j_{m-1}}}r^{\scriptscriptstyle\min}_{S_{j_{m-1}}, S_{j_{m-2}}} \cdots r^{\scriptscriptstyle\min}_{S_{j_1}, S_{\ell}}, 
  \quad j=\ell+1, \ldots, k,    
\end{align}   where $P_{\mathrm{sub}}[\ell, j]=\{\ell, j_1, \cdots, j_{m-1}, j\}$ represents a subsequence of the path $[\ell,j]$.  
Then the hitting probability  
$ 
h^{{\scriptscriptstyle\min}}_{S_k,S_\ell} \ge c^{\scriptscriptstyle\min}_{k,\ell}. $ 
\end{theorem} 
\begin{proof}
   Equations \eqref{eq:recursion}, \eqref{eq:expression-sum2} and \eqref{eq:expression-sum} are equivalent because the recursive formula \eqref{eq:recursion} can be expanded into the non-recursive formula \eqref{eq:expression-sum2} and then written as \eqref{eq:expression-sum} using subsequence notation. By Corollary \ref{cor:recursive} and Equation \eqref{eq:recursion}, we  obtain \( h^{{\scriptscriptstyle\min}}_{S_j,S_\ell} \ge c^{\scriptscriptstyle\min}_{j,\ell} \).
\end{proof}

The above theorem provides a non-recursive formula for calculating the best lower bound coefficient $c^{\scriptscriptstyle\min}_{k,\ell}$.    
However, its computation is considered computationally intractable because the number of subsequences is $(k-\ell)!$. Formulas \eqref{equ:LowerCoeff-c} and \eqref{equ:LowerCoeff-cl} give two explicit expressions for quickly estimating the lower bound coefficients. However, these lower bounds are loose for fitness landscapes with shortcuts \cite{he2025drift}, and therefore they are not applicable to non-level-based fitness functions. To address this issue, we propose new non-recursive formulas based on the concepts of paths and segments.   

The following theorem proves that the best coefficient $c^{\scriptscriptstyle\min}_{k,\ell}$ is no less than the product of the conditional probability $r(X_i,S_{[\ell,i)})$ staying on the complete path $[\ell, i)$. The new formula \eqref{eq:expression-prod} greatly simplifies the computation of  coefficient $c_{k,\ell}$ because it only uses $k-\ell$ conditional transition probabilities.

\begin{theorem}\label{the:ExplicitExpression1}
Given an elitist EA, a level partition $(S_0, \ldots, S_K)$, and two levels $S_\ell,  S_k: 1 \le \ell < k \le K$,  for any $\ell <j \le k$,  let coefficients $c_{\ell,\ell}=1$ and 
\begin{align}
\label{eq:expression-prod}
c_{j,\ell}= \prod_{i \in (\ell,j]} r^{\scriptscriptstyle\min}_{S_i,S_{ [\ell ,i)}}, \quad j=\ell+1, \cdots, k. 
\end{align}
 Then  
$   c^{\scriptscriptstyle\min}_{k,\ell} \ge c_{k, \ell}. $  
\end{theorem}

\begin{proof}   
We prove the statement that $c^{\scriptscriptstyle\min}_{j,\ell} \ge c_{j, \ell}$ for $ \ell<j\le k$ by induction on $j$. 
For $j=\ell+1$, the statement is true because
$$c^{\scriptscriptstyle\min}_{\ell+1,\ell} =  c_{\ell+1, \ell}=r^{\scriptscriptstyle\min}_{S_{\ell+1}, S_\ell}.$$ 

We make the inductive assumption that $c^{\scriptscriptstyle\min}_{i,\ell} \ge c_{i, \ell}$ for $i = \ell+1, \cdots, j$, and want to prove $c^{\scriptscriptstyle\min}_{j+1,\ell} \ge c_{j+1, \ell}$ by induction. Note that the value of the coefficient $c_{j,\ell}$ decreases as $j$ increases.
From \eqref{eq:recursion} and the inductive assumption, we have 
\begin{align*} 
c^{\scriptscriptstyle\min}_{j+1,\ell}&=r^{\scriptscriptstyle\min}_{S_{j+1}, S_\ell} +  \sum^{j}_{i=\ell+1}  r^{\scriptscriptstyle\min}_{S_{j+1}, S_i} c^{\scriptscriptstyle\min}_{i,\ell}
\ge r^{{\scriptscriptstyle\min}}_{S_{j+1}, S_\ell} + \sum^{j}_{i=\ell+1}r^{{\scriptscriptstyle\min}}_{S_{j+1}, S_i}  c_{i,\ell} \\
 & 
\ge  r^{{\scriptscriptstyle\min}}_{S_{j+1}, S_\ell}  
  + r^{{\scriptscriptstyle\min}}_{S_{j+1}, S_{[\ell+1,j]}} c_{j,\ell} \ge \left(r^{{\scriptscriptstyle\min}}_{S_{j+1}, S_\ell}  
  + r^{{\scriptscriptstyle\min}}_{S_{j+1}, S_{[\ell+1,j]}}\right) c_{j,\ell} \\
  &=      r^{{\scriptscriptstyle\min}}_{S_{j+1}, S_{[\ell,j]}} c_{j,\ell} =c_{j+1,\ell}.  
\end{align*} 
We get $c^{\scriptscriptstyle\min}_{j+1,\ell} \ge c_{j+1, \ell}$ and then complete the induction proof.  
\end{proof}

The following corollary replaces the complete path $[\ell,k] $ in Theorem \ref{the:ExplicitExpression1} by any  path $P[\ell,k]$.

\begin{corollary}\label{cor:ExplicitExpression1}
Given an elitist EA, a level partition $(S_0, \ldots, S_K)$ and two levels $S_k, S_\ell$ (where$1 \le \ell < k \le K$), let $P[\ell,k]$ be a path from  $S_k$ to $S_\ell$. 
Then the following holds:
\begin{align}
\label{eq:expression-prod-path}
 c^{\scriptscriptstyle\min}_{k,\ell}  
 &\ge    \prod_{i \in P(\ell,k]} r^{\scriptscriptstyle\min}_{S_i,S_{ P[\ell ,i)}}. 
\end{align}   
\end{corollary} 

In some cases, we consider multiple paths because the hitting probability aggregated over multiple paths exceeds the hitting probability over a single path.  
To handle multiple paths, we introduce a new technique based on decomposing a path into segments. For example, Figure~\ref{fig:Subset}.P5-b displays multiple paths from the empty knapsack \( L_{(0,0;0)} \) to the local optima \( L_{(0,1; n/2 - 2)} \). A path can be decomposed into three segments, as shown below.
\begin{align}
\overbrace{L_{(0,0;0)} \to \cdots \to}^{\mathrm{1st}} \overbrace{L_{(0,0;i)} \to L_{(0,1;i)}}^{\mathrm{2nd}} \overbrace{\to \cdots \to L_{(0,1;n/2-2)}}^{\mathrm{3rd}}, & \quad i = 0, \ldots, n/2-2. \label{equ:multiple-paths}
\end{align} 
The following theorem gives a general case with three segments and shows that the hitting probability can be estimated by combining these segments. %The theorem holds for any number of paths from $S_k$ to $S_\ell$. 
Although Theorem \ref{the:ExplicitExpression2} focuses on combinations of three segments, it can be generalized to combinations of any number of segments.

\begin{theorem}\label{the:ExplicitExpression2}
Given an elitist EA, a level partition $(S_0, \ldots, S_K)$ and two levels $S_k, S_\ell$ (where $1 \le \ell < k \le K$), consider $m$ paths $P_i[\ell,k]$ (where $m\ge 1$ and $i=1, \cdots, m$) from $S_k$ to $S_\ell$ and each path is decomposed into three segments: $P_i[\ell,\mu_i]$,  $P_i[\mu_i, \lambda_i]$, and $P_i[\lambda_i, k]$. 
\begin{align*}
 \overbrace{S_k \to \cdots \to}^{\mathrm{1st}} \overbrace{S_{\mu_i} \to  \cdots \to S_{\lambda_i}}^{\mathrm{2nd}} \overbrace{\to \cdots \to S_{\ell}}^{\mathrm{3rd}}, & \quad i =0, \ldots, m.
\end{align*} 
Let $P_{i\_{\mathrm{sub}}}[\mu_i,k]= k  \to j_m \to \cdots   \to j_1 \to  \mu_i $ denote a subsequence of segment $P_i[\mu_i,k]$, $P_{i\_\mathrm{sub}}[\lambda_i,\mu_i]= \mu_i \to j'_m \to   \cdots   \to j'_1\to  \lambda_i $ denote a subsequence of segment $P_i[\lambda_i,\mu_i]$,  and $P_{i\_\mathrm{sub}}[\ell, \lambda_i]= \lambda_i \to j''_m \to \to \cdots \to j''_1 \to  \ell $ denote a subsequence of segment  $P_i[\ell,\lambda_i]$. 
%Let \(P_{i\_\mathrm{sub}}[\ell,k]\) denote the concatenation    $ P_{i\_\mathrm{sub}}[\ell,\lambda_i]^\frown P_{i\_\mathrm{sub}}[\lambda_i,\mu_i]^\frown P_{i\_\mathrm{sub}}[\mu_i,k]$ which is a subsequence from $S_k$ to $S_\ell$.   
Assign the coefficient 
\begin{align} 
  c_{k,\ell}  
&=\sum^m_{i=1} \left(\sum_{P_{i\_\mathrm{sub}}[\mu_i,k] \subseteq P_i[\mu_i,k]}  \, r^{\scriptscriptstyle\min}_{S_{k}, S_{j_{m-1}}}r^{\scriptscriptstyle\min}_{S_{j_{m-1}}, S_{j_{m-2}}} \cdots r^{\scriptscriptstyle\min}_{S_{j_1}, S_{\mu_i}} \right) \nonumber\\ 
&\quad \times \left(\sum_{P_{i\_\mathrm{sub}}[\lambda_i,\mu_i] \subseteq P_i[\lambda_i,\mu_i]}  \, r^{\scriptscriptstyle\min}_{S_{\mu_i}, S_{j'_{m-1}}}r^{\scriptscriptstyle\min}_{S_{j'_{m-1}}, S_{j'_{m-2}}} \cdots r^{\scriptscriptstyle\min}_{S_{j'_1}, S_{\lambda_i}} \right)\nonumber\\
&\quad \times\left( \sum_{P_{i\_\mathrm{sub}}[\ell, \lambda_i] \subseteq P_i[\ell,\lambda_i]}  \, r^{\scriptscriptstyle\min}_{S_{\lambda_i}, S_{j''_{m-1}}}r^{\scriptscriptstyle\min}_{S_{j''_{m-1}}, S_{j''_{m-2}}} \cdots r^{\scriptscriptstyle\min}_{S_{j''_1}, S_{\ell}}\right),
 \label{eq:expression-segments}
\end{align} 
If for every \(i = 1,\ldots,m\), each subsequence (concatenation)
\[
P_{i_{\mathrm{sub}}}[\ell,\lambda_i]^\frown  
P_{i_{\mathrm{sub}}}[\lambda_i,\mu_i]^\frown  
P_{i_{\mathrm{sub}}}[\mu_i,k]
\]  from \(S_k\) to \(S_\ell\) 
is distinct from all others, then  
\(c^{\min}_{k,\ell} \geq c_{k,\ell}\).
\end{theorem}

\begin{proof} 
In the formula \eqref{eq:expression-segments}, the first inner sum enumerates the subsequences $P_{i\_\mathrm{sub}}[\mu_i,k]$ from \( S_k \) to \( S_{\mu_i} \), the second inner sum enumerates the  subsequences $P_{i\_\mathrm{sub}}[\lambda_i,\mu_i]$ from \( S_{\mu_i} \) to \( S_{\lambda_i} \), and the third inner sum enumerates the subsequences $P_{i\_\mathrm{sub}}[\ell, \lambda_i]$ from \( S_{\lambda_i} \) to \( S_\ell \). The combination of them forms subsequences from \( S_k \) to \( S_\ell \). The theorem condition states that these subsequences are distinct from each other. By Theorem \ref{the:ExplicitExpression0}, we obtain  $c^{\scriptscriptstyle\min}_{k,\ell} \ge c_{k, \ell}$.
\end{proof}

The proof above relies on the fact that the partial sum of subsequences does not exceed the total sum of  subsequences. The key point of this argument is that there is no duplicate subsequences from $S_k$ to $S_\ell$. One criterion for this distinction is that each path $P_i[\ell, k]$  contains one segment whose subsequences are not shared with any other path. In other words, this segment intersects with no more than one vertex from any other path. 
For example, in Figure ~\ref{fig:Subset}.P5-b, the second segment of the $i$th path in \eqref{equ:multiple-paths}, $L_{(0,0;i)} \to L_{(0,1;i)}$, shares only a single vertex with any other path.
%However, directly using subsequences of multiple paths from \( S_k \) to \( S_\ell \) does not guarantee that these subsequences are distinct from each other, because any path from \( S_k \) to \( S_\ell \) contains the repeated  subsequence $k \to \ell$.  Therefore, segments are needed to handle multiple paths.

In the theorem above, the  direct computation of coefficients $c_{k,\mu_i}$, $c_{\mu_i,\lambda_i}$, and $c_{\lambda_i,\ell}$ is intractable since the number of subsequences can grow exponentially.  However, it  can be simplified by Corollary \ref{cor:ExplicitExpression1}. This leads to the following computation \eqref{eq:expression-segments-2} using $\sum^m_{i=1}(k-\mu_i) (\mu_i-\lambda_i) (\lambda_i-\ell)$  conditional probabilities. 

\begin{corollary} \label{cor:ExplicitExpression2}
Under the same condition as Theorem \ref{the:ExplicitExpression2}, it holds:  
\begin{align} 
& c^{\scriptscriptstyle\min}_{k,\ell} \ge  \sum^m_{i=1} %\left(
\prod_{j\in P_i(\mu_i,k]} r^{\scriptscriptstyle\min}_{S_j,S_{P_i [\ell ,j)}} %\right) \left(
\prod_{j\in P_i(\lambda_i,\mu_i]} r^{\scriptscriptstyle\min}_{S_j,S_{ P_i[\ell ,j)}}%\right) \left(
\prod_{j\in P_i(\ell,\lambda_i]} r^{\scriptscriptstyle\min}_{S_j,S_{ P_i[\ell ,j)}}%\right)
. 
 \label{eq:expression-segments-2}
\end{align}   
\end{corollary}

%Based on Theorem \ref{the:HittingTime} and Corollary \ref{cor:ExplicitExpression1}, we derive a new explicit formula that quickly computes a lower bound on the mean hitting time as follows:
%\begin{align}
% \frac{1}{p^{\max}_{S_k,S_{[0,k-1]}}} + \sum_{\ell =1}^{k-1} \frac{ c_{k,\ell} }{ p^{\max}_{S_\ell,S_{[0,\ell-1]}} }, &\quad \mbox{ where } c_{k,\ell} =\prod_{i\in P(\ell, k]} r^{\scriptscriptstyle\min}_{S_i,S_{ [\ell ,i-1]}}.
%\end{align}

%Similarly based on Corollary \ref{cor:ExplicitExpression2}, we can derive an explicit formulation that incorporates multiple paths and segments.
%Compared to the previously established lower bounds \citep{sudholt2012new, doerr2024lower, he2025drift}, these new lower bounds show improved tightness, especially for non-level-based fitness functions, as shown in the next section.

%%%%%%%%%%%%%%%%%%%%%%%%%%%%%%%%%%%
\subsection{Product Inequalities} 
\label{sec:Product}
The lower bound coefficient is related to product calculation. The following lemma establishes three  product inequalities that are very useful in practical estimation of lower bounds.

\begin{lemma} \label{lem:product} For any positive constant $C$,  the following inequalities hold.
    \begin{align}
        & \lim_{n \to \infty}  \prod^n_{i =1} \frac{1}{1+  \frac{C}{n-1}} 
  \ge e^{-C}.\label{equ:product1}
  \\ 
  &\lim_{n \to \infty}  \prod^n_{i =1} \frac{1}{1+   \frac{C}{i!} }   
  \ge e^{-C(e-1)}.\label{equ:product2}
  \end{align}

For  any positive sequences $\{a_i, i=1, \ldots, n\}$ and $\{b_i, i=1, \ldots, n$), it holds
  \begin{align} 
  &\prod^n_{i =1} \frac{1}{1+a_i+b_i}   \ge \prod^n_{i = 1} \frac{1}{1+a_i} \prod^n_{i =1}\frac{1}{1+b_i}.
  \label{equ:product3}
    \end{align}
\end{lemma}

\begin{proof}
The first two inequality is proven as follows.
\begin{align*}
    &\lim_{n \to \infty} \prod^n_{i =1} \frac{1}{1+  \frac{C}{n-1}} 
  \ge \lim_{n \to \infty}  \left(1-\frac{C}{n-1}\right)^{n-1}  = e^{-C}. 
\end{align*}

The second inequality is proven as follows.
\begin{align*}
 & \lim_{n \to \infty} \prod^n_{i=1}  \frac{1}{1+  \frac{C}{i!}}  = \lim_{n \to \infty}  \exp\left({-\sum^n_{i=1} \ln\left(1+  \frac{C}{i!}\right)}\right) \\
 &\ge \lim_{n \to \infty} \exp\left( {-\sum^n_{i=1} \frac{C}{i!}}\right) =e^{-C(e-1)}.
\end{align*}

The third inequality is derived from $(1+a)(1+b) \ge 1+a+b$ for any $a \ge 0, b \ge 0$. 
\end{proof}

%%%%%%%%%%%%%%%%%%%%%%%%%%%%%%%%%%%%%%%%%%%%%
\section{Applications of the Subset Fitness Level Method}
\label{sec:Applications} 
Using the six knapsack instances P1 to P6, this section shows that the subset method yields tight lower bounds for non‑level‑based functions.

\subsection{Instance P1} 
Figure~\ref{fig:Subset}.P1 shows that, for  P1, the local optimal set is $L_{(0,1;1)}$. We select the subset  \(S' \) with a direct path from $L_{(0,0;0)}$ to this local optimum:
$
 L_{(0,0;0)} \to L_{(0,1;1)}.
$
The subset \( S' \) is partitioned into fitness levels as follows, ordered by decreasing fitness value. 
\[ 
\begin{array}{lll}
S_1 = L_{(0,1;1)},    \\
S_2 = L_{(0,0;0)}.
\end{array} 
\]  

By Theorem \ref{the:HittingTime}, the mean hitting time from the empty knapsack $L_{(0,0;0)}$ to $S_0$ (where $S_0=S \setminus S'$) is lower-bounded by
\begin{align}
    & m_{(0,0;0), S_0} \ge  \frac{  h_{(0,0;0),(0,1;1)}}{ p_{(0,1;1),(0,1;1)^+}}   \ge \frac{  h_{(0,0;0),(0,1;1)}}{ p_{(0,1;1),(1,0;0) \cup (0,0;[{n}/{2}+1,n-2])}}.
\label{instance1-1} 
\end{align} 

The transition from $L_{(0,1;1)}$ to $L_{(1,0;0)}$ happens only if bits $b_1, b_2$ and the one-valued bit in $[b_3,b_n]$ are flipped. This gives a transition probability 
\begin{align} 
 p_{(0,1;1), (1,0;0)}  &\le  \frac{1}{n^3}.
\label{instance1-2}
\end{align}

The transition from $L_{(0,1;1)}$ to $L_{(0,0;[n/2+1,n-2])}$ happens only if bit $b_2$ and at least $n/2$ of $n-3$ zero-valued bits in $[b_3,b_n]$ are flipped.  This gives a transition probability    
\begin{align}  
  p_{(0,1;1), (0,0;[{n}/{2}+1,n-2])} &\le \frac{1}{n} \binom{n-3}{{n}/{2}}\left(\frac{1}{n}\right)^{{n}/{2}}  
  \le   \frac{1}{ n ({n}/{2})!  } \label{instance1-3}.  
\end{align}
Therefore, by combining \eqref{instance1-2} and \eqref{instance1-3}, we get the transition probability
\begin{align}  p_ {(0,1;1), (1,0,0) \cup (0,0;[{n}/{2}+1,n-2])}
\le \frac{1}{n^3} + \frac{1}{n({n}/{2})! } = O\left(\frac{1}{n^3}\right). \label{instance1-4}
\end{align}

The transition from $L_{(0,0;0)}$ to $L_{(0,1;1)}$ happens if bit $b_2$ and one of $n-2$ zero-valued bits in $[b_3,b_n]$ are flipped, and other bits are unchanged. According to Theorem \ref{the:ExplicitExpression1}, the hitting probability $h_{(0,0;0), (0,1;1)}$ is lower-bounded by
\begin{align}  
  h_{(0,0;0),(0,1;1)} \ge p_{(0,0;0),(0,1;1)}
  \ge   \frac{1}{n} \binom{n-2}{1} \frac{1}{n} \left(1-\frac{1}{n}\right)^{n-2}.  \label{instance1-5} 
\end{align} 

By inserting \eqref{instance1-4} and \eqref{instance1-5} to \eqref{instance1-1}, we derive the following lower bound on the mean hitting time from the empty knapsack $L_{(0,0;0)}$ to \( S_0 \): 
\begin{align*} 
m_{(0,0;0), S_0}  =  \Omega( n^{3})\; \Omega(n^{-1})    =\Omega(n^2). 
\end{align*}  
Since \( S_0 \supset S_{\mathrm{opt}} \), it follows that the mean hitting time from the empty knapsack to the optimal set is $
 \Omega(n^2).$
%%%%%%%%%%%%%%%%%%%%%%%%%

\subsection{Instance P2} 
Figure~\ref{fig:Subset}.P2 shows that for  P2, the local optimal set is $L_{(0,1;0)}$. We select the subset  \(S' \) with a direct path from $L_{(0,0;0)}$ to this local optimum:
$ 
 L_{(0,0;0)} \to  L_{(0,1;0)}.
 $ 
The subset \( S' \) is partitioned into fitness levels as follows, ordered by decreasing fitness value. 
\[ 
\begin{array}{lll}
S_1 = L_{(0,1;0)},    \\
S_2 = L_{(0,0;0)}. 
\end{array} 
\] 
 
According to Theorem \ref{the:HittingTime}, the mean hitting time from the empty knapsack $L_{(0,0;0)}$ to $S_0$ (where $S_0=S \setminus S'$) is lower-bounded by
\begin{align}
   & m_{(0,0;0), S_0} \ge  \frac{h_{(0,0;0),(0,1;0)}}{p_{ (0,1;0),(0,1; 0)^+}}  = \frac{h_{(0,0;0),(0,1;0)}}{p_{ (0,1;0),(0,0;[\frac{n}{2}+1, n-3])}}.\label{instance2-1}
\end{align}  
 
The transition from $L_{(0,1;0)}$ to $L_{(0,0;[n/2+1, n-3])}$ happens only if bit $b_2$ and at least $n/2+1$\ of $n-2$ zero-valued bits in $[b_3,  b_n]$ are flipped. This gives a transition probability 
\begin{align} 
   p_{ (0,1;0),(0,0;[{n}/{2}+1, n-3])} \le & \frac{1}{n} \binom{n-2}{{n}/{2}+1}\left(\frac{1}{n}\right)^{{n}/{2}+1} 
   \le  \frac{1}{n({n}/{2}+1)!}. \label{instance2-2}
\end{align} 
 
The transition from $L_{(0,0;0)}$ to $L_{(0,1;0)}$ happens if $b_2$ is flipped and other bits are unchanged.  According to Theorem \ref{the:ExplicitExpression1}, the hitting probability 
\begin{align}
   h_{(0,0;0),(0,1;0)} \ge r_{(0,0;0),(0,1;0)}  \ge p_{(0,0;0),(0,1;0)}  
     \ge   \frac{1}{n} \left(1-\frac{1}{n}\right)^{n-1} . \label{instance2-3}
\end{align}

Substituting \eqref{instance2-2} and \eqref{instance2-3} into \eqref{instance2-1}, we derive the following lower bound on the mean hitting time from the empty knapsack $L_{(0,0;0)}$ to \( S_0 \): 
\begin{equation}  
m_{(0,0;0), S_0} = \Omega\left(\left(\frac{n}{2}+1\right)! \right) .  
\end{equation}
Since \( S_0 \supset S_{\mathrm{opt}} \), it follows that the mean hitting time from the empty knapsack to the optimal set is $\Omega\left(({n}/{2}+1)! \right).$

%%%%%%%%%%%%%%%%%%%%%%%%
\subsection{Instance P3}
Figure~\ref{fig:Subset}.P3 shows that for P3, the two local optima sets are $L_{(0,1;1)}$ and $L_{(0,0;n-2)}$. 
We select the subset  \(S' \) with the following path towards the local optimum $L_{(0,0;n-2)}$:
$L_{(0,0;0)} \to  L_{(0,0;1)} \to \cdots \to L_{(0,0;n-2)}.$  The subset \( S' \) is subsequently partitioned into fitness levels as follows, ordered by decreasing fitness value. 
\begin{equation*}  
\begin{array}{llll}
      S_\ell  &= L_{(0,0; i)}, & i = n/2-2,\ldots,0, & \ell = n/2-1-i.
\end{array}
\end{equation*} 

According to Theorem \ref{the:HittingTime}, the mean hitting time from the empty knapsack $L_{(0,0;0)}$ to $S_0$ (where $S_0=S \setminus S'$) is lower-bounded by
\begin{align} 
  m_{(0,0;0), S_0}
    \ge \frac{h_{(0,0;0), (0,0;n-2)}}{p_{(0,0;n-2), (0,0;n-2)^+}}  =\frac{h_{(0,0;0), (0,0;n-2)}}{p_{(0,0;n-2), (1,0;0)}}.\label{instance3-1}
\end{align}   

The transition from $L_{(0,0;n-2)}$ to $L_{(1,0;0)}$ happens only if bit $b_1$ and all bits in $[b_3,b_n]$ are flipped.  This gives a transition probability 
\begin{align} 
    p_{(0,0;n-2), (1,0;0)}
    \le \left(\frac{1}{n}\right)^{n-1}. \label{instance3-2}
\end{align}

We estimate the hitting probability $h_{(0,0;0),(0,0;n-2)}$  by  Theorem \ref{the:ExplicitExpression1}.  
\begin{align}
  &     h_{(0,0;0),(0,0;n-2)} \ge  \prod_{j=1}^{n-2} r_{(0,0;j),(0,0;[j+1,n-2])}  =  \prod_{j=1}^{n-2}\frac{p_{(0,0;j),(0,0;[j+1,n-2])}}{p_{(0,0;0),(0,0;j)^+}} \nonumber \\
 &\ge \frac{p_{(0,0;0),(0,0;[1,n-2])}}{1} \prod_{j=1}^{n-3}\frac{p_{(0,0;j),(0,0;[j+1,n-2])}}{p_{(0,0;j),(0,0;[j+1,n-2])}+p_{(0,0;j),(1,0;0)}+p_{(0,0;j),(0,1;[0,1])}} \nonumber \\
  &  \ge p_{(0,0;0),(0,0;[1,n-2])}  \prod_{j=1}^{n-3}\frac{1}{1+\frac{p_{(0,0;j),(1,0;0)}+p_{(0,0;j),(0,1;[0,1])}}{p_{(0,0;j),(0,0;[j+1,n-2]))}}}.\label{instance3-3}
\end{align}

The transition from $L_{(0,0;0)}$ to $L_{(0,0;[1,n-2])}$ happens if one of $n-2$ zero-valued bits in $[b_3, b_n]$ is flipped and other bits are unchanged. This gives a transition probability 
\begin{equation} 
     p_{(0,0;0),(0,0;[1,n-2]))}\ge \binom{n-2}{1}\frac{1}{n}\left(1-\frac{1}{n}\right)^{n-1} \ge \frac{n-2}{en}.  \label{instance3-4}
\end{equation}

For $j=1, \ldots, n-1$, the transition from $L_{(0,0;j)}$ to $L_{(0,0;[j+1,n-2])}$ happens if one of $n-2-j$ zero-valued bits in $[b_3, b_n]$ is flipped and other bits are unchanged. This gives a transition probability 
\begin{equation} 
     p_{(0,0;j),(0,0;[j+1,n-2]))}\ge \binom{n-2-j}{1}\frac{1}{n}\left(1-\frac{1}{n}\right)^{n-1} \ge \frac{n-2-j}{en}. \label{instance3-5}
\end{equation}

The transition from $L_{(0,0;j)}$ to $L_{(1,0;0)}$ happens only if bit $b_1$ and all $j$ one-valued bits in $[b_3,b_n]$ are flipped. This gives a transition probability 
\begin{equation} 
    p_{(0,0;j),(1,0;0)} \le \left(\frac{1}{n}\right)^{1+j} . \label{instance3-6}
\end{equation}  

The transition from $L_{(0,0;j)}$ to $L_{(0,1;[0,1])}$ happens only if  bit $b_2$ and at least $j-1$ of $j$ one-valued bits in $[b_3, b_n]$ are flipped.
\begin{equation} 
     p_{(0,0;j),(0,1;[0,1])}\le  
      \frac{1}{n}\binom{j}{j-1}\left(\frac{1}{n}\right)^{j-1}. \label{instance3-7}
\end{equation}

By inserting \eqref{instance3-4}, \eqref{instance3-5}, \eqref{instance3-6}, and \eqref{instance3-7} into \eqref{instance3-3}, we obtain  
\begin{align}
 h_{(0,0;0),(0,0;n-2)} 
   & \ge\frac{n-2}{en}   \prod_{j=1}^{n-3}\frac{1}{1+ \left({\left(\frac{1}{n}\right)^{1+j}+\binom{j}{j-1}\left(\frac{1}{n}\right)^{j}}\right)\div{\frac{n-2-j}{en}}}\nonumber\\
     & \ge  \Omega(1)  \prod_{j=1}^{n-3}\frac{1}{1+   \left(\frac{1}{n}\right)^{j}   } \prod_{j=1}^{n-3}\frac{1}{1+    \frac{1}{(j-1)!}} =\Omega(1). \label{instance3-8}
\end{align}
The constant bound $\Omega(1)$ follows from Lemma~\ref{lem:product}.

Finally, by substituting \eqref{instance3-2} and \eqref{instance3-8} into \eqref{instance3-1},  we derive the following lower bound on the mean hitting time from the empty knapsack $L_{(0,0;0)}$ to \( S_0 \): 
$$m_{(0,0;0), S_0} = \Omega(n^{n-1}).$$ 

Since \( S_0 \supset S_{\mathrm{opt}} \), it follows that the mean hitting time from the empty knapsack  to the optimal set is
$ \Omega(n^{n-1}).$
 
%%%%%%%%%%%%%%%%%%%%%%%%
\subsection{Instance P4} 
Figure~\ref{fig:Subset}.P4 shows that for Instance P4,  the local optima sets are $L_{(1,0;0)}$ and $L_{(0,0;n-2)}$. 
we select the subset  \(S' \) with a direct path from $L_{(0,0;0)}$ to the local optimum $L_{(1,0;0)}$:
$
 L_{(0,0;0)} \to L_{(1,0;0)}.
$
The subset \( S' \) is  partitioned into fitness levels as follows, ordered by decreasing fitness value. 
\begin{align*} 
     S_1 = L_{(1,0;0)},    \\
    S_2 =  L_{(0,0;0)}.  
\end{align*}

According to Theorem \ref{the:HittingTime}, the mean hitting time from the empty knapsack $L_{(0,0;0)}$ to $S_0$ is lower-bounded by
\begin{align}
 m_{(0,0;0),S_0}   
    \ge \frac{h_{(0,0;0), (1,0;0) }}{ p_{(1,0;0) ,(1,0;0)^+} } =\frac{h_{(0,0;0), (1,0;0) }}{ p_{(1,0;0) ,(0,1;[{n}/{2}-2, {3n}/{4}-2]) \cup (0,0;n-2)} }.\label{instance4-1}
\end{align} 

The transition from $L_{(1,0;0)}$ to $L_{(0,1;[{n/}{2}-2,{3n}/{4}-2])}$ happens only if bits $b_1, b_2$ and at least $n/2-2$ of $n-2$ zero-valued bits in $[b_3,b_n]$ are flipped. This gives a transition probability 
\begin{align}
    p{_{(1,0;0),(0,1;[{n}/{2}-2, {3n}/{4}-2])}}\le& \left(\frac{1}{n}\right)^2 \binom{n-2}{{n}/{2}-2} \left(\frac{1}{n}\right)^{{n}/{2}-2} \le   \frac{1}{ n^2({n}/{2}-2)! }. \label{instance4-2}
\end{align} 
The transition from $L_{(1,0;0)}$ to $L_{(0,0;n-2)}$ happens only if bit $b_1$ and all $n-2$ bits in $[b_3,b_n]$ are flipped. This gives a transition probability 
\begin{align}
    p_{(1,0;0),(0,0;n-2) }\le&   \left(\frac{1}{n}\right)^{n-1}  . \label{instance4-3}
\end{align} 

 By combining \eqref{instance4-2} and \eqref{instance4-3}, we get an upper bound on the transition probability from \( L_{(1,0;0)} \) to \( L_{(0,0;[n/2-2,3n/4-2])} \cup L_{(0,0;n-2)}\) as follows   
\begin{align}
p_{(1,0;0),(0,1;[{n}/{2}-1, {3n}/{4}-2]) \cup (0,0;n-2) } &\le \frac{1}{ n^2({n}/{2}-2)! }+\left(\frac{1}{n}\right)^{n-1} \nonumber\\
&=O\left(\frac{1}{ n^2({n}/{2}-2)! }\right).\label{instance4-4}
\end{align}

We estimate the hitting probability $h_{(0,0;0), (1,0;0)}$  by Theorem \ref{the:ExplicitExpression1}.  The transition from $L_{(0,0;0)}$ to $L_{(1,0;0)}$ happens  if bit $b_1$ is flipped and other bits are unchanged. 
The hitting probability 
\begin{align}
    h_{(0,0;0), (1,0;0)}  \ge  p_{(0,0;0),(1,0;0) }  
    \ge  \frac{1}{n} \left(1-\frac{1}{n}\right)^{n-1}.\label{instance4-5}
\end{align}
 
By inserting \eqref{instance4-4} and \eqref{instance4-5} into \eqref{instance4-1},  we derive the following lower bound on the mean hitting time from the empty knapsack $L_{(0,0;0)}$ to \( S_0 \): 
\begin{equation}
    m_{(0,0;0),S_0} = \Omega\left(n^2 \left(\frac{n}{2}-2\right)! \right)  \Omega\left(\frac{1}{n}\right) = \Omega\left(n\left(\frac{n}{2}-2\right)! \right). 
\end{equation}
 Since \( S_0 \supset S_{\mathrm{opt}} \), it follows that the mean hitting time from the empty knapsack to the optimal set is $ \Omega\left(n ({n}/{2}-2)! \right).$ 
%%%%%%%%%%%%%%%%%%%%%%%%
\subsection{Instance P5}  

Figure~\ref{fig:Subset}.P5-b illustrates that, for Instance P5, the set of local optima is given by \( L_{(0,1; n/2 - 2)} \). We consider multiple paths from the empty knapsack state \( L_{(0,0;0)} \) to the local optimum \( L_{(0,1; n/2 - 2)} \) as follows.
Each path is decomposed into three segments:
\[
\overbrace{L_{(0,0;0)} \to \cdots \to}^{\text{1st}} 
\overbrace{L_{(0,0;i)} \to L_{(0,1;i)}}^{\text{2nd}} 
\overbrace{\to \cdots \to L_{(0,1; n/2 - 2)}}^{\text{3rd}}, 
\quad i = 0, \ldots, n/2 - 2.
\]

We define the subset \( S' \) to consist of all vertices lying along these paths. The subset \( S' \) is partitioned into fitness levels as follows, ordered by decreasing fitness value. 
\begin{equation*}  
\begin{array}{llll}
      &S_{\ell}  = L_{(0,1; i)}, & i=n/2-2, \cdots, 0, & \ell =n/2-1-i\\  
      &S_{\ell}= L_{(0,0; i)}, & i=n/2-2, \cdots, 0, & \ell = n-2-i.
\end{array} 
\end{equation*} 

According to Theorem \ref{the:HittingTime}, the mean hitting time from the empty knapsack $L_{(0,0;0)}$ to $S_0$ (where $S_0=S\setminus S'$) is lower-bounded by
\begin{align}
 m_{(0,0;0),S_0}    
    \ge \frac{h_{(0,0;0), (0,1; {n}/{2}-2) }}{p_{(0,1; {n}/{2}-2),(0,1; {n}/{2}-2)^+}} =\frac{h_{(0,0;0), (0,1; {n}/{2}-2) }}{p_{(0,1; {n}/{2}-2),(0,0;[n-3,n-2])}}.\label{instance5-1}
\end{align} 

The transition from $L_{(0,1;n/2-2)}$ to $L_{(0,0;[n-3,n-2])}$ happens only if bit $b_2$ and at least $n/2-1$ bits of $n/2$ zero-valued bits in $[b_3,b_n]$ are flipped. 
\begin{align*}
    p_{(0,1; {n}/{2}-2),(0,0;[n-3,n-2])}\le& \frac{1}{n} \binom{{n}/{2}}{{n}/{2}-1} \left(\frac{1}{n}\right)^{\frac{n}{2}-1} \le   \frac{1}{ n( {n}/{2}-1)! }.%\label{instance5-2}
\end{align*}
Then we get a lower bound on the mean hitting time as 
\begin{align}
 m_{(0,0;0),S_0}    
    \ge \frac{h_{(0,0;0), (0,1; {n}/{2}-2) }}{p_{(0,1; {n}/{2}-2),(0,1; {n}/{2}-2)^+}} \ge   n( {n}/{2}-1)! \times h_{(0,0;0), (0,1; {n}/{2}-2) } \label{instance5-3}
\end{align} 

We prove the hitting probability $h_{(0,0;0),(0,1;n/2-2)}=\Omega(1)$ by Corollary \ref{cor:ExplicitExpression2}.
According to Corollary \ref{cor:ExplicitExpression2}, we obtain the lower bound coefficients composed of the lower bound coefficients on the three segments as follows:
\begin{align}
\label{equ:P5-z1}
c_{(0,0;0),(0,1;n/2-2)} = \sum^{{n}/{2}-2}_{i=0} c_{(0,0;0),(0,0;i)} \times c_{(0,0;i),(0,1;i)} \times c_{(0,1;i),(0,1;n/2-2)}.
\end{align}

For the first segment, $L_{(0,0;0)} \to \cdots \to L_{(0,0;i)}$ (where $i \in [0,n/2-2]$), we have 
\begin{align*}
 &c_{(0,0;0), (0,0;i) }  =     \prod_{j=0}^{i-1} r_{(0,0;j),(0,0;[j+1,i])} =  \prod_{j=0}^{i-1} \frac{p_{(0,0;j),(0,0;[j+1,i])} }{ p_{(0,0;j),(0,0;j)^+} }\nonumber\\
   & =    \prod_{j=0}^{i-1} \frac{p_{(0,0;j),(0,0;[j+1,i])} }{ p_{(0,1;j),(0,1;[j+1,i])}+ p_{(0,0;j),(0,0;[i+1,n/2-2])}+ p_{(0,0;j),(0,1;[0,n/2])} } \nonumber \\ 
   & =     \prod_{j=0}^{i-1} \frac{1 }{ 1+\frac{p_{(0,0;j),(0,0;[i+1,n/2-2])} +p_{(0,0;j),(0,1;[0,n/2])}}{p_{(0,0;j),(0,0;[j+1,i])} } }.
\end{align*} 

Since the transition probabilities
\begin{align*}
& p_{(0,0;j),(0,0;[i+1,n/2-2])} \le \binom{n-2-j}{i+1-j} \left(\frac{1}{n}\right)^{i+1-j}, \\
    &p_{(0,0;j),0,1;[0,n/2])} \le \frac{1}{n}, \\
    &p_{(0,0;j),(0,0;[j+1,i])} \ge \binom{n-2-j}{1} \frac{1}{n} \left(1-\frac{1}{n}\right)^{n-1},
\end{align*}
we get \begin{align}
 &c_{(0,0;0), (0,0;i) }    \ge     \prod_{j=0}^{i-1} \frac{1 }{ 1+ \frac{e}{(i+1-j)!}+ \frac{e }{n-2-j}} \nonumber\\
 &\ge \prod_{j=0}^{n/2-3} \frac{1 }{ 1+ \frac{e}{(n/2-1-j)!}+ \frac{e}{n/2+1}} =\Omega(1).
 \label{equ:P5-z2}
\end{align}

For the second segment, $L_{(0,0;i)} \to L_{(0,1;i)}$ (where $i \in [0,n/2-2]$),  we have
\begin{align}
c_{(0,0;i),(0,1;i)} = r_{(0,0;i),(0,1;i)} > p_{(0,0;i),(0,1;i)} = \frac{1}{n} \left(1-\frac{1}{n}\right)^{n-1} = \Omega(n^{-1}). \label{equ:P5-z3}
\end{align}

For the third segment, $L_{(0,1;i)} \to \cdots \to L_{(0,1;n/2-2)}$ (where $i \in [0,n/2-2]$), we have 
\begin{align*}
 &c_{(0,1;0), (0,1;i) }  =      \prod_{j=0}^{i-1} r_{(0,1;j),(0,1;[j+1,i])} =     \prod_{j=0}^{i-1} \frac{p_{(0,1;j),(0,1;[j+1,i])} }{ p_{(0,1;j),(0,1;j)^+}  }  \nonumber\\
   & =       \prod_{j=0}^{i-1} \frac{p_{(0,1;j),(0,1;[j+1,i])} }{ p_{(0,1;j),(0,1;[j+1,i])}+p_{(0,1;j),(0,1;[i+1,n/2-2])} +p_{(0,1;j), (0,0; [n/2-1+j, n-2]}}   \\
   &=  \prod_{j=0}^{i-1} \frac{1 }{ 1+\frac{p_{(0,1;j),(0,1;[i+1,n/2-2])} +p_{(0,1;j), (0,0; [n/2-1+j, n-2]} }{p_{(0,1;j),(0,1;[j+1,i])} } }. 
\end{align*} 

Since the transition probabilities
\begin{align*}
& p_{(0,1;j),(0,1;[i+1,n/2-2])} \le \binom{n-2-j}{i+1-j} \left(\frac{1}{n}\right)^{i+1-j}. \\ 
    &p_{(0,1;j), (0,0; [n/2-1+j, n-2]} \le \frac{1}{n} \binom{n-2-j}{n/2-1} \left(\frac{1}{n}\right)^{n/2-1}. \\
    &p_{(0,1;j),(0,1;[j+1,i])} \ge \binom{n-2-j}{1} \frac{1}{n} \left(1-\frac{1}{n}\right)^{n-1}.
\end{align*}
we get \begin{align}
 c_{(0,1;0), (0,1;i) }   
   &\ge     \prod_{j=0}^{i-1} \frac{1 }{ 1+ \frac{e}{(i+1-j)!}+ \frac{e}{(n/2-2-j) (n/2-1)!} } \nonumber\\ 
   &\ge \prod_{j=0}^{n/2-3} \frac{1 }{ 1+ \frac{e}{(n/2-1-j)!} + \frac{e}{(n/2-2-j) (n/2-1)!}} =\Omega(1). \label{equ:P5-z4}
\end{align}

Substituting \eqref{equ:P5-z2}, \eqref{equ:P5-z3}, and \eqref{equ:P5-z4} into \eqref{equ:P5-z1}, we obtain a lower bound on the hitting probability $h_{(0,0;0),(0,1;n/2-2)}$ as
\[
 c_{(0,0;0),(0,1;n/2-2)} = \sum_{i=0}^{{n}/{2}-2} \Omega(1) \times \Omega(n^{-1}) \times \Omega(1) = \Omega(1).
\]
 We establish the desired lower bound:  \( h_{(0,0;0),\,(0,1;n/2-2)} = \Omega(1) \).  By substituting this lower bound into \eqref{instance5-3}, we derive the following lower bound on the mean hitting time from the empty knapsack $L_{(0,0;0)}$ to \( S_0 \): 
\[
m_{(0,0;0),S_0}   \ge \Omega\left(n\left(\frac{n}{2} - 1\right)!\right).  
\]
 Since \( S_0 \supset S_{\mathrm{opt}} \), it follows that the mean hitting time from the empty knapsack to the optimal set is
$ \Omega\left(n\left( {n}/{2} - 1\right)!\right).$

\subsection{Instance P6 and Linear Functions} 
The fitness function for instance P6 is linear. It is well established that the upper bound for linear functions is \( O(n \log n) \) \citep{he2001drift,droste2002analysis}. In this paper, we establish a matching lower bound of \( \Omega(n \ln n) \) using Theorem \ref{the:ExplicitExpression1}. According to the theory of the easiest fitness functions, ONEMAX is considered the simplest among all linear functions \citep{he2015easiest}. Therefore, it suffices to prove that the lower bound for ONEMAX is \( \Omega(n \ln n) \).   

ONEMAX is given by $f(x) =|x|:=x_1+\cdots+x_n$, which is a level-based function. We partition the entire search space $S$ into fitness levels as follows. 
\begin{align*}
    S_{\ell}= 
      \{x; |x|=n-\ell \},   \quad \ell=0, \ldots, n.
\end{align*}

According to Theorem \ref{the:HittingTime}, the mean hitting time from the empty knapsack $S_n$ to $S_0$ is lower-bounded by
\begin{align} 
  m_{S_{n}, S_0}
    \ge \frac{1}{p_{S_{n},S_{[0,n-1]}}}+\sum_{\ell =1}^{n-1}\frac{  h_{S_{n},S_\ell} }{ p_{S_\ell,S_{[0,\ell -1]}}} .\label{instance6-1}
\end{align}

For $\ell=1, \ldots, n$,  the transition from $S_\ell$ to $S_{[0,\ell-1]}$ happens only if one of $\ell$ zero-valued bits is flipped. This gives a transition probability 
\begin{align} 
p_{S_\ell,S_{[0,\ell -1]}} \le  \frac{\ell}{n}.  \label{instance6-2}
\end{align}

The hitting probability $h_{S_{n},S_\ell}$ is estimated using  Theorem \ref{the:ExplicitExpression1}.  
\begin{align}
    h_{S_n,S_1} \ge   \prod_{i=\ell+1}^n r_{S_i,S_{ [\ell ,i-1]}} = \prod_{i=\ell+1}^n 
    \frac{1}{1+\frac{p_{S_i,S_{ [0 ,\ell-1]}}}{p_{S_i,S_{ [\ell ,i-1]}}}}.   \label{instance6-3}
\end{align}

For $i =\ell+1, \ldots, n$, a transition from $S_i$ to $S_{[0,\ell-1]}$ occurs only if at least $i-\ell+1$ bits among the $i$ zero-valued bits are flipped. This gives a transition probability 
\begin{align}
p_{S_i,S_{[0,\ell-1]}} \le \binom{i}{i-\ell+1}\left(\frac{1}{n}\right)^{i-\ell+1}. \label{instance6-4}
\end{align}
A transition from $S_i$ to $S_{[\ell,i-1]}$ occurs if one of the $i$ zero-valued bits is flipped while the other bits remain unchanged. This gives a transition probability 
\begin{align}
p_{S_i,S_{[\ell,i-1]}} \ge \frac{i}{n} \left(1-\frac{1}{n}\right)^{n-1} \ge \frac{i}{ne}.
\label{instance6-5}
\end{align}

By substituting \eqref{instance6-4} and \eqref{instance6-5} into \eqref{instance6-3}, we obtain the hitting probability:
\begin{align}
    h_{S_n,S_1} \ge  \prod_{i=\ell+1}^n
    \frac{1}{1+ \binom{i}{i-\ell+1}\left(\frac{1}{n}\right)^{i-\ell+1}  \div {\frac{i}{ne}}} \ge   \prod_{i=2}^n 
    \frac{1}{1+ \frac{e}{(i-1)!}} =\Omega(1). \label{instance6-6}
\end{align}
The constant bound \(\Omega(1)\) is derived from Lemma~\ref{lem:product}.

By substituting \eqref{instance6-2} and \eqref{instance6-6} into \eqref{instance6-1}, we derive a lower bound as below:
\begin{align} 
  m_{S_{n}, S_0}
    \ge 1 + \Omega(1) \sum_{\ell = 1}^{n-1} \frac{n}{\ell} = \Omega(n \log n).
\end{align} 
This lower bound is true for all linear functions because ONEMAX is the easiest. 

Table~\ref{tab:results} summarizes the analysis results for six instances. As can be seen from the table  (instances P1 to P5), the lower bound obtained by partitioning a subset of non-optimal solutions is much tighter than the lower bound obtained by partitioning the entire search space.

 \begin{table}[ht]
     \centering
     \begin{tabular}{c|c|c}
     \toprule
      instance & subset &entire search space\\
      \midrule
         P1 &   $\Omega\left(n^2 \right)$   & \( O(n \log n) \) \\ \midrule
         P2  & $\Omega\left(({n}/{2}+1)! \right)$ & \( O(n \log n) \)\\  \midrule
         P3 & $ \Omega(n^{n-1})$ &\( O(n \log n) \)\\ \midrule
         P4 & $ \Omega\left(n ({n}/{2}-2)! \right)$ & \( O(n \log n) \)\\ \midrule
         P5 & $ \Omega\left(n \left( {n}/{2} - 1\right)!\right)$ & \( O(n \log n) \) \\ \midrule
          P6 & $\Omega(n \log n)$  &\( O(n \log n) \)\\
          \bottomrule
     \end{tabular}
     \caption{Comparison of the lower bounds obtained by partitioning a subset of non-optimal solutions with the lower bounds obtained by partitioning the entire search space.}
     \label{tab:results}
 \end{table} 
%%%%%%%%%%%%%%%%%%%%%%%%%%%
 \section{Conclusions}
\label{sec:Conclusion} 
This study identifies a shortcoming of the classical fitness level method: it does not yield tight lower bounds on hitting times when the fitness function does not naturally induce level-based structure. To overcome this issue, we propose the subset fitness level method, which focuses on a selected subset of non‑optimal solutions and partitions it into fitness levels from which lower bound coefficients can be derived by drift analysis. We also provide new explicit expressions, grounded in the concepts of paths and segments, that enable fast computation of these coefficients directly from conditional transitional probabilities. Empirical evaluation on six knapsack instances demonstrates that our method consistently yields tight lower bounds, proving its efficacy for analyzing fitness functions without level-based structures.

A remaining task is to prove that the above lower bounds match the corresponding upper bounds \eqref{equ:Upper-bound}. 
In addition to the knapsack problem, future work will explore potential applications extended to other combinatorial optimization problems such as the arc routing problem \citep{tong2024evaluating} or  EAs for continuous optimization \citep{chen2021average}.

%%%%%%%%%%%%%%%%%%%%%%%%

 \end{document}